\documentclass[letterpaper]{article} % DO NOT CHANGE THIS
\usepackage{aaai2027}  % DO NOT CHANGE THIS
\usepackage{times}  % DO NOT CHANGE THIS
\usepackage{helvet}  % DO NOT CHANGE THIS
\usepackage{courier}  % DO NOT CHANGE THIS
\usepackage[hyphens]{url}  % DO NOT CHANGE THIS
\usepackage{graphicx} % DO NOT CHANGE THIS
\usepackage{natbib}  % DO NOT CHANGE THIS AND DO NOT ADD ANY OPTIONS TO IT
\usepackage{caption} % DO NOT CHANGE THIS AND DO NOT ADD ANY OPTIONS TO IT
\usepackage{algorithm}
\usepackage{algorithmic}
\usepackage{amsmath}
\usepackage{amsfonts}
\usepackage{booktabs} 
\usepackage{multirow}  
\usepackage{colortbl}  
\usepackage{xcolor}
\usepackage{tcolorbox}
\tcbuselibrary{skins, breakable}
\usepackage{newfloat}
\usepackage{listings}
\DeclareCaptionStyle{ruled}{labelfont=normalfont,labelsep=colon,strut=off} % DO NOT CHANGE THIS
\floatstyle{ruled}
\newfloat{listing}{tb}{lst}{}
\floatname{listing}{Listing}
\newcommand{\sysname}{CERA-MoA}
\title{\sysname: Co-Evolving Routing Mechanisms with Continually Learning LLM Agents}
\author{
	Jiaxuan Jiang\textsuperscript{1, 3},
	Liyuan He\textsuperscript{2, 3},
	Zhixuan Fang\textsuperscript{1, 3, \thanks{Corresponding author}}
}
\affiliations{
	\textsuperscript{1}IIIS, Tsinghua University, Beijing, China\\
	\textsuperscript{2}School of Artificial Intelligence, Shanghai Jiao Tong University, Shanghai, China\\
	\textsuperscript{3}Shanghai Qi Zhi Institute, Shanghai, China
}

\nocopyright
\usepackage{bibentry}
\begin{document}

\maketitle

\begin{abstract}
Current Mixture-of-Agents (MoA) paradigms generally treat query routing and agent fine-tuning as separate processes, limiting their ability to respond to evolving agent capabilities. This disconnect prevents routing strategies from adapting to evolving agent capabilities during post-training and prevents agents from achieving synergistic data-driven specialization. To resolve this, we introduce \sysname~(Co-Evolving Router with continually learning Agents for Mixture-of-Agents), an iterative reinforcement learning framework where the dynamic router and independent agent policies co-evolve. We design a predictive familiarity estimator that leverages mid-layer hidden states to evaluate semantic competence among agents, avoiding the overhead of full rollouts. Based on these familiarity scores, a cumulative-threshold adaptive routing mechanism dynamically activates a tailored minimal agent subset, achieving a trade-off between task performance and efficiency. By proactively allocating targeted training samples to agents based on their evolving competence, \sysname~promotes capability differentiation. Extensive experiments across various domains demonstrate that \sysname~outperforms state-of-the-art static-agent routing and fix-workflow fine-tuning baselines.
\end{abstract}


\section{Introduction}
Although large language models (LLMs) have demonstrated remarkable reasoning capabilities, standard post-training methods frequently encounter generalization bottlenecks when dealing with increasingly complex and multifaceted problem domains \cite{hong2024metagpt, ye2025mas}. To transcend the limitations of single monolithic models, the Mixture-of-Agents (MoA) paradigm has emerged as a promising solution. By combining multiple agents, this paradigm aims to solve diverse tasks more effectively by leveraging the complementary strengths of different agents \cite{wang2024mixtureofagentsenhanceslargelanguage}. Such collaborative frameworks have the potential to substantially expand the performance boundaries of LLMs on complex tasks.

Despite this potential, the current development of multi-agent systems primarily bifurcates into two directions. The first direction focuses on optimizing orchestration mechanisms among fixed-capability agents, which is typically achieved through debate frameworks to refine reasoning consensus \cite{chan2023chateval,estornell2024multi,yi2025debate,hu2025multiagentdebatellmjudges,fan2025imadintelligentmultiagentdebate,qiao2026epistemicgainaleatoriccost}, or through dynamic agent selection for efficient query allocation \cite{10.1145/3589334.3645420, yue2025masrouter, lee2026confidence, poon2026online, wang2026icl, xue2026r2routernewparadigmllm, wang2026routemoa}. The second direction explores training and fine-tuning agents, but typically operates within predetermined multi-agent workflow architectures \cite{park2025maporlmultiagentpostcotrainingcollaborative, motwani2025maltimprovingreasoningmultiagent, marti2025, xue2026comascoevolvingmultiagentsystems, zhao2026strongermasmultiagentreinforcementlearning, wang2026maspojointpromptoptimization}. Although these approaches effectively enhance collective performance, they inherently decouple the routing strategy from the continual learning dynamics of agents. 

This decoupling exposes a critical research gap in contemporary multi-agent paradigms. First, the capabilities of individual agents continually evolve during the post-training phase. However, existing routing mechanisms are not designed to adapt to such capability shifts. Second, current post-training pipelines typically rely on manually partitioned datasets, lacking a dynamic sample allocation mechanism. Without routing specific training queries to agents based on their competence, the system fails to automatically induce distinct, targeted expertise. Consequently, the disconnect between routing strategies and the continual learning of agents prevents the system from achieving synergistic capability specialization, leaving individual agents acting as generalists rather than domain experts.

To bridge this critical disconnect, we formulate {Co-Evolving Router with continually learning Agents for Mixture-of-Agents (\sysname)}, an iterative closed-loop paradigm that integrates agent learning with dynamic routing optimization. Rather than treating routing and agent adaptation in isolation, \sysname~not only adapts to the progressively shifting capabilities of individual agents, but also dynamically allocates tailored training queries to explicitly induce skill specialization. At the core of our system is a predictive familiarity estimator, which directly extracts the intermediate hidden states of LLMs to quantify how well an input query aligns with each agent's learned expertise. This design provides a dynamic evaluation of relative competence prior to text generation, reducing the heavy computational overhead of external evaluators or full rollout generations. Utilizing these familiarity scores, we further introduce a cumulative-threshold adaptive routing strategy. Rather than relying on a fixed top-$k$ agent allocation, this mechanism dynamically selects a varying number of agents based on the estimated competence coverage of the agent population. By activating the smallest score-ranked subset of agents whose cumulative familiarity exceeds the threshold, our strategy achieves a fine-grained trade-off between task performance and computational cost.

The contributions of this work are summarized as follows:
\begin{itemize}
    \item We introduce a sample-level Mixture-of-Agents framework that enables continual reinforcement learning of LLM agents through adaptive query routing. This co-evolutionary system simultaneously adapts the routing mechanism to shifting agent capabilities and dynamically allocates training queries thereby encouraging distinct problem-solving specialization.
    \item We design a predictive familiarity estimator for efficient semantic-level competence evaluation, together with a cumulative-threshold adaptive routing mechanism to dynamically balance performance and computational cost based on the estimated competence.
    \item We conducted extensive experiments demonstrating that our framework outperforms static-agent routing and fix-workflow fine-tuning baselines. Empirical results highlight performance gains across diverse problem domains.
\end{itemize}

\section{Related Work}

Multi-agent systems (MAS) have emerged as a powerful paradigm to extend the reasoning boundaries of large language models (LLMs) by leveraging collective intelligence \cite{zhao2024electoral, ye2025mas}. To facilitate effective collaboration, standard MAS architectures typically organize models through structured interaction protocols, such as multi-agent debate \cite{liang2023encouraging, estornell2024multi}, majority voting \cite{chen2023reconcile, taubenfeld2025confidence}, or mixtures of independent agents \cite{wang2023fusing, xie-etal-2025-rmoa, li2026ofa}. Unlike token-level Mixture-of-Experts (MoE) architectures that route internal hidden representations across specialized sub-networks \cite{oldfield2024multilinearmixtureexpertsscalable, lv2025couplingexpertsroutersmixtureofexperts, zhuang2025ld}, MAS operates at the sample and semantic level, requiring high-level coordination among autonomous models. To optimize the collective efficacy of these collaborative systems, current research methodologies generally bifurcate into two directions: {\textbf{orchestration optimization}}, which focuses on dynamic interaction protocols among fixed-capability agents, and {\textbf{agent fine-tuning}}, which actively trains individual agent policies within predefined workflows.

\paragraph{Orchestration Optimization for Multi-Agent Systems.}
To optimize collaboration structure, existing literature widely investigates dynamic orchestration and query routing. Approaches range from multi-arm bandits \cite{10.1145/3589334.3645420, poon2026online} and knapsacks within budgets \cite{wang2025mixllm, xue2026r2routernewparadigmllm} to agent diversity maximization \cite{xie-etal-2025-rmoa}, lightweight evaluation scorers \cite{yue2025masrouter, wang2026icl, wang2026routemoa}, and confidence-guided stepwise routing \cite{lee2026confidence, wang2026orchestrating}. Recently, studies have also explored leveraging LLMs directly as self-orchestrators \cite{dang2025multiagentcollaborationevolvingorchestration, ke2026masorchestraunderstandingimprovingmultiagent}, topology graph generators \cite{zhang2026evorouteexperiencedrivenselfroutingllm, li2026ofa}, or meta-thinkers \cite{zhu2026recognizeorchestratorentropydynamics}, while methods like AgentDropout \cite{wang-etal-2025-agentdropout} prune redundant communication nodes to reduce overhead. Despite effectively optimizing orchestration and reducing costs, these techniques generally assume that candidate models remain static during the orchestration training phase. Consequently, they lack an integrated mechanism to realign query allocation as individual models actively evolve and specialize. \sysname~bridges this gap by co-evolving a predictive familiarity estimator that captures dynamic relative competence alongside agent policy updates, ensuring dynamic adaptation to shifting agent expertise.

\paragraph{Agent Fine-tuning for Multi-Agent Systems.}
Beyond static interactions, recent work actively fine-tunes agents to enhance individual and collaborative reasoning using reinforcement learning, preference optimization, and supervised learning. The methods explore test-time self-verification \cite{pmlr-v267-lee25ab}, reasoning chain refinement \cite{puerto2025fine}, self-reflection cycles \cite{zhao2025learningreasonexternalrewards}, and reflection interaction optimization \cite{yuan2025reinforce}. Within collaborative setups, fine-tuning is driven by rule-based verifiers \cite{park2025maporlmultiagentpostcotrainingcollaborative}, tree-structured sampling \cite{motwani2025maltimprovingreasoningmultiagent, zhao2026strongermasmultiagentreinforcementlearning}, LLM-as-a-judge interactions \cite{xue2026comascoevolvingmultiagentsystems}, and end-to-end multi-agent reinforcement learning \cite{wang2026maspojointpromptoptimization}. However, existing fine-tuning pipelines predominantly optimize agent policies within predefined workflow architectures. Furthermore, they mainly rely on manually partitioned or uniform training datasets, which keep data allocation separate from real-time learning dynamics. Without capability-aware query routing during training, existing systems lack an explicit mechanism to guide agents toward complementary specialization, leaving agents to act as homogeneous generalists. \sysname~addresses this gap by integrating an iterative routing mechanism directly into the training loop, proactively allocating semantic queries to agents based on their evolving competence to explicitly foster domain specialization.

\section{\sysname}
\begin{figure*}[h]
    \centering
    \includegraphics[width=\linewidth]{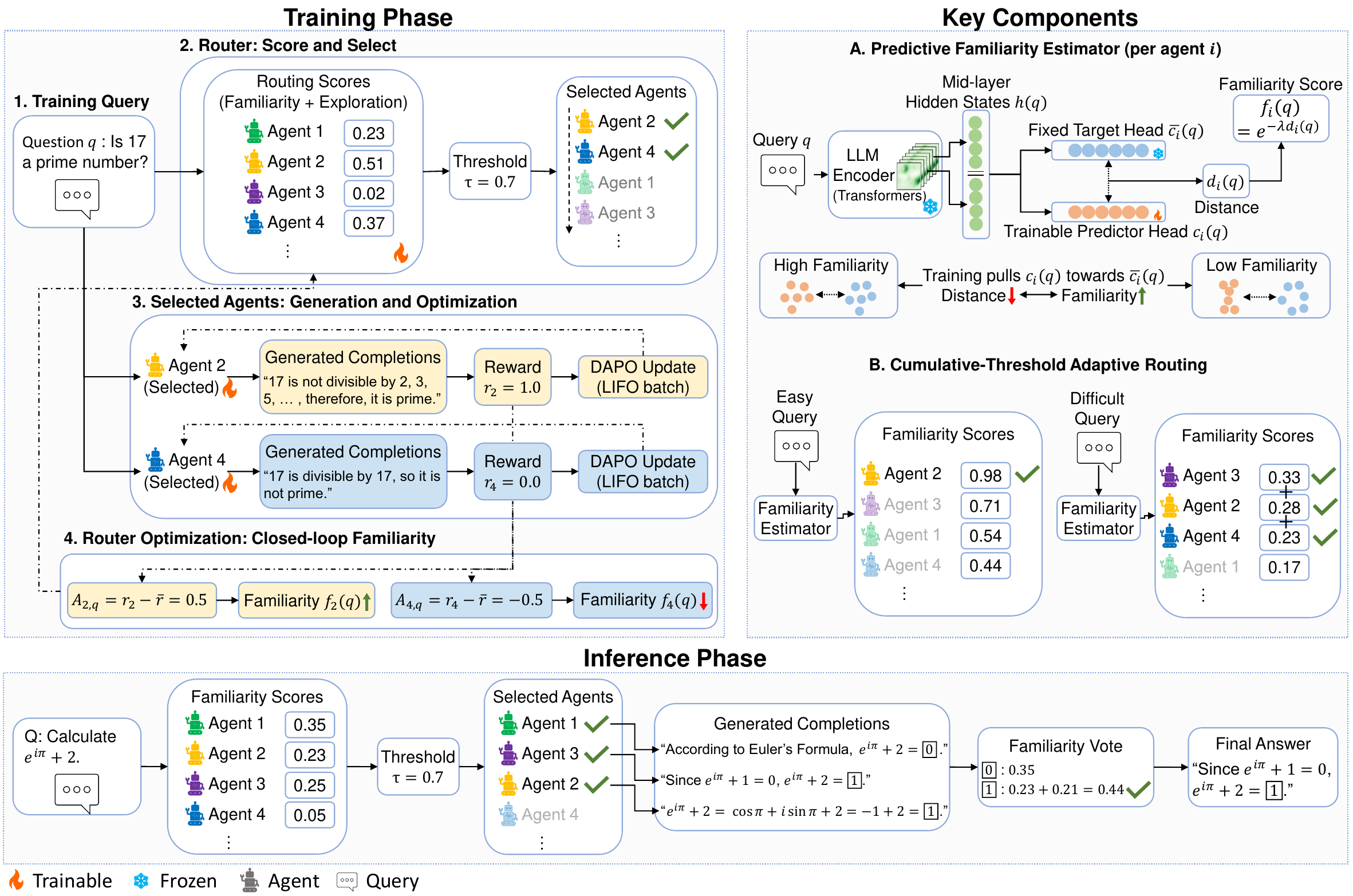}
    \caption{Overview of \sysname.}
    \label{fig:overall_framework}
\end{figure*}
\subsection{Overview}
As illustrated in Figure~\ref{fig:overall_framework}, we formulate \sysname~(\underline{C}o-\underline{E}volving \underline{R}outer with continually learning \underline{A}gents for Mixture-of-Agents) as a collaborative framework $\mathcal{M} = \left(\mathcal{D}_\psi, \left\{\mathcal{A}_{\theta_i}\right\}_{i=1}^N, \mathcal{S}\right)$, integrating a parameterized router $\mathcal{D}_\psi$, a population of continually learning agent policies $\left\{\mathcal{A}_{\theta_i}\right\}_{i=1}^N$, and a voting-based aggregator $\mathcal{S}$. Our primary design objective is to establish a closed-loop co-evolutionary process for the router and the agents: adaptive query allocation routes targeted training samples to specific agents to drive domain specialization, while the router synchronously updates its evaluation parameters to accurately track these continually shifting agent capabilities.

\paragraph{Training Phase.}
For an incoming query, the router $\mathcal{D}_\psi$ computes a familiarity score that measures relative agent competence and allocates the sample to a tailored subset of agents. Selected agents independently generate completions and optimize their policies through reinforcement learning, driven by task-specific rewards. Concurrently, the router evaluates the relative advantages of agent performance to update its familiarity estimator. This iterative feedback loop naturally encourages specialization: agents receive problems aligned with their potential, evolving from homogeneous generalists into domain specialists, while the router synchronously tracks their real-time expertise.

\paragraph{Inference Phase.}
During inference, $\mathcal{D}_\psi$ routes the query to the minimal subset of competent agents based on learned familiarity scores. Each activated agent generates a single response. For tasks with deterministic solutions, the aggregator $\mathcal{S}$ executes familiarity-weighted majority voting:
\begin{equation}
\hat{y} = \arg\max_y \sum_{i \text{ s.t. } \text{ans}(a_i)=y} f_i(q),
\end{equation}
selecting the final completion from the agent exhibiting the highest familiarity score $f_i(q)$ within the winning consensus group. For open-ended tasks, $\mathcal{S}$ directly outputs the completion of the most familiar agent.

\subsection{Predictive Familiarity Estimator}
To estimate agent-query compatibility prior to generation without relying on costly external verifiers, we propose a predictive familiarity estimator. It consists of a trainable predictor head $g_i$ and a randomly initialized and permanently frozen target head $\bar{g}_i$ for each agent $i$. The estimator projects query features extracted from the router's frozen backbone into a reference space. Importantly, the backbone itself is not updated by the familiarity objective; only the predictor head is trained. Thus, the intermediate hidden states $h(q)$ should be understood as fixed semantic features produced by the shared backbone, while what evolves during training is the learned mapping from these features to the familiarity score space. As agent policies co-evolve and the routed training distribution shifts, the predictor head adapts to the changing competence landscape over this fixed feature space.

Query $q$ is first passed through the router's base language model. Its semantic representation $h(q)$ is then computed by concatenating intermediate hidden states from the model. Intermediate hidden states have been shown to capture semantic information complementary to final-layer representations, whose are more directly tied to next-token prediction \cite{skean2025layerlayeruncoveringhidden}. Let $\operatorname{hidden}_{l}(q)$ denote the hidden state at layer $l$ of a total depth $L$. We define the semantic representation as
\begin{equation}
h(q) = \operatorname{concat}\left(\operatorname{hidden}_{\lfloor\frac{L}{2}\rfloor}(q) \middle\| \operatorname{hidden}_{\lfloor\frac{3L}{4}\rfloor}(q)\right).
\end{equation}
The semantic representation is projected into two distinct embeddings $c_i(q) = g_i(h(q))$ and $\bar{c}_i(q) = \bar{g}_i(h(q))$. The target heads remain frozen and are initialized orthogonally across different agents to provide a stable reference space while ensuring initial diversity among the agent population. The normalized Euclidean distance between these two projections is computed to quantify the head distance $d_i(q)$ according to
\begin{equation}
d_i(q) = \left\| \frac{c_i(q)}{\|c_i(q)\|_2} - \frac{\bar{c}_i(q)}{\|\bar{c}_i(q)\|_2} \right\|_2.
\end{equation}
The familiarity score is then derived via exponential decay:
\begin{equation}
f_i(q) = \exp(-\lambda d_i(q)),
\end{equation}
where $\lambda$ is a temperature hyperparameter controlling the sensitivity of the score to the projection distance. A smaller distance therefore yields a higher familiarity score. Rather than regressing an absolute reward, the estimator converts reward supervision into a relative push-and-pull signal around a fixed anchor. The frozen target head does not encode a semantic prototype or competence label; it simply provides a stable reference point. Positive relative advantages pull the predictor toward this anchor, while negative advantages push it away by a margin. In this way, the familiarity score becomes a reward-aligned geometric compatibility signal in a fixed reference space.

\subsection{Cumulative-Threshold Adaptive Routing}
To transcend fixed top-$k$ routing constraints, we introduce a familiarity-aware, cumulative-threshold allocation mechanism that balances reasoning efficacy against multi-agent inference overhead by activating the minimal agent subset required to reach a target competence threshold.

During training, routing scores incorporate exploration bonuses alongside familiarity scores to prevent starvation, as well as historical entropy to encourage policy diversity:
\begin{equation}
s_i(q) = f_i(q) + c_{ucb} \sqrt{\frac{\log T}{n_i + 1}} + w_{ent} \overline{H}_i,
\end{equation}
where $T$ is the number of routed queries, $n_i$ is the number of training samples allocated to agent $i$, and $\overline{H}_i$ represents the agent's historical mean generation entropy. Let $(j)$ denote the index of the agent ranked $j$-th in descending order by the routing score $s_i(q)$. Crucially, while the routing score $s_i(q)$ determines the {priority} of agent selection to foster exploration during training, the cutoff for subset activation is strictly evaluated against familiarity scores. Thus, the router activates the smallest top-ranked subset of agents whose cumulative familiarity score $f_{(j)}(q)$ satisfies:
\begin{equation}
\min_k \sum_{j=1}^k f_{(j)}(q) \ge \tau,
\end{equation}
where $\tau$ is the predefined cumulative threshold. If the total sum falls below $\tau$, the query is allocated to the entire population. At inference time, the exploration terms are deactivated ($c_{ucb}=w_{ent}=0$), strictly routing queries to the most competent specialists while preserving efficiency.

\subsection{Joint Optimization of Router and Agents}
The engine driving \sysname~is a closed-loop reinforcement learning protocol that bridges agent policy optimization with synchronous router updates. This training mechanism explicitly promotes capability differentiation: agents that successfully solve a query acquire higher familiarity scores within that specific semantic space. Meanwhile, they become more likely to be allocated semantically similar queries in the future, establishing a learning loop that naturally induces deep domain specialization.

\paragraph{Agent Optimization.} 
Selected agents store allocated queries in a last-in-first-out (LIFO) buffer. Upon forming a training batch, agent $i$ samples $G$ completions per query and updates its policy using Dynamic Sampling Policy Optimization (DAPO) \cite{yu2025dapoopensourcellmreinforcement} combined with sequence-level importance sampling from Group Sequence Policy Optimization (GSPO) \cite{zheng2025groupsequencepolicyoptimization}. 

Let $R\left(q,a_{i,q}^{(g)}\right)$ denote the reward assigned to the $g$-th completion of question $q$ by agent $i$. The normalized advantage for the $g$-th completion is
\begin{equation}
\hat{A}_{i,q}^{(g)}
=
\frac{
R\left(q,a_{i,q}^{(g)}\right)-\mu_{i,q}
}{
\sigma_{i,q}+10^{-4}
},
\end{equation}
where $\mu_{i,q}$ and $\sigma_{i,q}$ are the empirical mean and standard deviation of the $G$ rewards. 

For a completion $a_{i,q}^{(g)}=(y_1,\ldots,y_S)$, the sequence-level importance ratio is averaged over valid completion tokens:
\begin{equation}
\rho_{i,q}^{(g)} = \exp\left( \frac{1}{S} \sum_{t=1}^{S} \log \frac{\pi_{\theta_i}(y_t\mid q,y_{<t})}{\pi_{\theta_i^{\mathrm{old}}}(y_t\mid q,y_{<t})} \right).
\end{equation}
Then we define the clipped surrogate as
\begin{equation}
s_{i,q}^{(g)}
=
-\min\left(
\rho_{i,q}^{(g)}\hat{A}_{i,q}^{(g)},
\operatorname{clip}\left(\rho_{i,q}^{(g)},1-\epsilon,1+\epsilon\right)\hat{A}_{i,q}^{(g)}
\right).
\end{equation}
The KL penalty is computed token-wise against the reference policy. Defining
\begin{equation}
x_t
=
\log \pi_{\mathrm{ref}}\left(y_t\mid q,y_{<t}\right)
-
\log \pi_{\theta_i}\left(y_t\mid q,y_{<t}\right),
\end{equation}
the token-level KL penalty then utilizes the estimator
\begin{equation}
D_{\mathrm{KL},t} = \exp(x_t) - x_t - 1.
\end{equation}
The loss for agent $i$ is normalized by the number of active completion tokens in the accumulated training batch:
\begin{equation}
\mathcal{L}_{\mathrm{agent}}(\theta_i)
=
\frac{
\sum_{q,g,t}
m_{q,g,t}
\left(s_{i,q}^{(g)}
+
\beta D_{\mathrm{KL},t}
\right)
}{
\sum_{q,g,t} m_{q,g,t}
},
\end{equation}
where $m_{q,g,t}$ masks padding tokens and $\beta$ acts as the regularization scaling hyperparameter.
\begin{table*}[t]
\centering
\small
\setlength{\tabcolsep}{5.5pt} % 调整列间距以适应双栏宽度
\begin{tabular}{l ccc ccc ccc c}
\toprule
\multirow{2}{*}{\textbf{Method}} & \multicolumn{3}{c}{\textbf{Math}} & \multicolumn{3}{c}{\textbf{Code}} & \multicolumn{2}{c}{\textbf{Instruction}} & \textbf{General} & \multirow{2}{*}{\textbf{Avg.}} \\
\cmidrule(lr){2-4} \cmidrule(lr){5-7} \cmidrule(lr){8-9} \cmidrule(lr){10-10}
& \textbf{GSM8k} & \textbf{MATH} & \textbf{DAPO-M} & \textbf{MBPP} & \textbf{Eurus} & \textbf{TACO} & \textbf{MAGPIE} & \textbf{RLVR-IF} & \textbf{BBH} & \\
\midrule
\textit{\textbf{Qwen3-4B}}          & 91.2 & 71.5 & 25.6 & 25.3 & 12.9 & 3.2 & 90.6 & 59.6 & 66.8 & 49.6 \\
\hspace{0.5em}+ ICL-Router         & 91.6 & 75.4 & 30.0 & 25.7 & 12.7 & 3.0 & 90.2 & 59.8 & 67.6 & 50.7 \\
\hspace{0.5em}+ LinUCB            & \underline{92.0} & 71.6 & 25.2 & 23.7 & 13.0 & 3.3 & \underline{91.2} & 59.4 & 70.6 & 50.0 \\
\hspace{0.5em}+ RouteMoA          & \textbf{93.5} & 75.6 & 28.8 & 32.7 & 15.1 & 3.5 & \textbf{92.8} & 62.0 & 74.2 & 53.1 \\
\hspace{0.5em}+ GSPO              & 91.4 & 76.1 & 35.6 & 59.9 & 27.3 & \underline{12.5} & {91.0} & 64.2 & 76.6 & 59.4 \\
\hspace{0.5em}+ MAPoRL            & {91.7} & 75.0 & \underline{38.4} & \textbf{64.2} & 23.8 & 6.4 & 89.0 & 59.8 & 76.4 & 58.3 \\
\hspace{0.5em}+ AT-GRPO      & 91.4 & \underline{77.6} & 38.0 & \underline{63.8} & \underline{29.8} & 12.0 & 90.4 & \underline{66.0} & \underline{79.8} & \underline{61.0} \\
\rowcolor{gray!10} \textbf{\hspace{0.5em}+ \sysname~(Ours)} & \textbf{93.5} & \textbf{79.6} & \textbf{41.8} & {60.3} & \textbf{31.3} & \textbf{13.4} & \textbf{92.8} & \textbf{72.4} & \textbf{83.4} & \textbf{63.2} \\
% \addlinespace[0.5em]
\midrule
% \multicolumn{11}{l}{\textit{\textbf{Base Model: Llama-3.2-3B-Instruct}}} \\
\textit{\textbf{Llama-3.2-3B-Instruct}}          & 62.9 & 34.6 & 10.4 & {41.2} & 4.9 & 0.8 & 66.4 & 41.0 & 39.8 & 33.6 \\
\hspace{0.5em}+ ICL-Router         & 64.8 & 40.8 & 10.8 & 43.6 & 7.2 & 1.3 & 68.6 & 46.0 & 47.0 & 36.7 \\
\hspace{0.5em}+ LinUCB            & 65.3 & 41.2 & 14.6 & 42.0 & 7.5 & 1.5 & 69.2 & 44.6 & 47.8 & 37.1 \\
\hspace{0.5em}+ RouteMoA          & 79.2 & 47.3 & \underline{16.2} & \textbf{49.0} & \underline{9.7} & 1.3 & \underline{79.0} & 50.8 & 57.4 & \underline{43.3} \\
\hspace{0.5em}+ GSPO              & \underline{81.5} & 47.1 & 14.8 & 28.0 & 7.8 & \underline{2.0} & {74.2} & 48.8 & 55.4 & 40.0 \\
\hspace{0.5em}+ MAPoRL            & 81.3 & \underline{49.1} & 14.8 & {40.9} & 6.8 & 1.1 & 73.6 & 43.2 & 57.8 & {41.0} \\
\hspace{0.5em}+ AT-GRPO      & 80.8 & 48.8 & \underline{16.2} & 25.3 & {7.9} & \textbf{2.5} & {74.2} & \underline{51.6} & \underline{59.4} & 40.7 \\
\rowcolor{gray!10} \textbf{\hspace{0.5em}+ \sysname~(Ours)} & \textbf{84.0} & \textbf{51.7} & \textbf{21.8} & \underline{44.7} & \textbf{11.3} & \underline{2.0} & \textbf{83.2} & \textbf{64.2} & \textbf{68.4} & \textbf{47.9} \\
% \addlinespace[0.5em]
\midrule
% \multicolumn{11}{l}{\textit{\textbf{Base Model: Phi-4-mini-Instruct}}} \\
\textit{\textbf{Phi-4-mini-Instruct}}          & 85.0 & 56.7 & 17.6 & 53.3 & 10.6 & 2.3 & 78.6 & 43.6 & 47.2 & 43.9 \\
\hspace{0.5em}+ ICL-Router         & 87.6 & 62.6 & 22.8 & 52.5 & 11.5 & 2.9 & 81.0 & 43.4 & 54.2 & 46.5 \\
\hspace{0.5em}+ LinUCB            & 86.9 & 63.0 & 20.6 & 51.4 & 14.9 & 3.0 & 82.0 & 43.4 & 54.2 & 46.6 \\
\hspace{0.5em}+ RouteMoA          & \textbf{91.5} & \underline{67.8} & \underline{24.2} & \textbf{60.7} & 15.5 & 4.7 & \underline{87.4} & 49.2 & 61.6 & 51.4 \\
\hspace{0.5em}+ GSPO              & 91.0 & 65.4 & 20.6 & 56.4 & \textbf{22.7} & \underline{8.3} & 87.2 & 56.2 & 73.4 & 53.5 \\
\hspace{0.5em}+ MAPoRL            & 89.6 & 67.1 & 24.0 & \underline{59.9} & 19.9 & 6.7 & 82.8 & 52.2 & 75.2 & 53.0 \\
\hspace{0.5em}+ AT-GRPO      & 89.8 & 66.9 & 23.6 & 54.9 & 22.1 & \textbf{8.4} & 87.0 & \underline{59.2} & \underline{75.4} & \underline{54.1} \\
\rowcolor{gray!10} \textbf{\hspace{0.5em}+ \sysname~(Ours)} & \underline{91.2} & \textbf{68.6} & \textbf{26.8} & 59.5 & \underline{22.2} & {7.7} & \textbf{87.6} & \textbf{61.0} & \textbf{76.0} & \textbf{55.6} \\
\midrule
\textit{\textbf{Heterogeneous pool}}          &  &  &  &  &  &  &  &  &  &  \\
\hspace{0.5em}+ ICL-Router         & 90.0 & 68.1 & 26.4 & 55.3 & 13.2 & 2.5 & 84.2 & 56.4 & 67.6 & 51.5 \\
\hspace{0.5em}+ LinUCB            & 91.7 & 70.9 & 26.0 & 46.3 & 13.0 & 3.1 & 90.2 & 57.2 & 69.6 & 52.0 \\
\hspace{0.5em}+ RouteMoA          & \underline{93.1} & 68.7 & 25.2 & \textbf{63.0} & 16.9 & 4.0 & 90.4 & 56.6 & 75.2 & 54.8 \\
\hspace{0.5em}+ MAPoRL            & 91.6 & 70.0 & 29.8 & 57.6 & \underline{23.9} & 9.0 & 88.8 & 61.2 & 76.6 & 56.5 \\
\hspace{0.5em}+ AT-GRPO      & \textbf{93.9} & \underline{76.2} & \underline{35.0} & \textbf{63.0} & \textbf{30.1} & \textbf{13.1} & \underline{91.8} & \underline{69.2} & \textbf{84.0} & \underline{61.8} \\
\rowcolor{gray!10} \textbf{\hspace{0.5em}+ \sysname~(Ours)} & \underline{93.1 } & \textbf{81.5} & \textbf{44.2} & \underline{59.5} & {23.8} & \underline{12.5} & \textbf{92.8} & \textbf{75.4} &\underline{83.8} & \textbf{63.0} \\
\bottomrule
\end{tabular}
\caption{Comparison of \sysname~with baselines on in-distribution (ID) tasks across different base models. The best is in \textbf{bold} and the second-best is \underline{underlined}.}
\label{tab:main_results}
\end{table*}
\paragraph{Router Optimization.}
The router updates by evaluating the relative performance of participating agents. For each query $q$, let $\mathcal{S}_q$ denote the set of agents activated in the current optimization step. We calculate the mean reward for agent $i$ over its sampled completions as
\begin{equation}
r_{i,q} = \frac{1}{G} \sum_{g=1}^G R\left(q, a_{i,q}^{(g)}\right).
\end{equation}
The relative advantage of agent $i$ is then computed by comparing its mean reward against the average performance across all participating agents:
\begin{equation}
A_{i,q} = r_{i,q} - \frac{1}{|\mathcal{S}_q|} \sum_{j \in \mathcal{S}_q} r_{j,q}.
\end{equation}
For solo-activated agents, we set $A_{i,q} = r_{i,q}$ to ensure a steady increase in familiarity score. To dynamically align routing decisions with evolving agent competencies, the router loss optimizes the predictor head $g_i$:
\begin{equation}
\mathcal{L}_{\text{router}}(g_i) = \sum_{q \in \mathcal{Q}} \begin{cases} A_{i,q} d_i(q), & \text{if } A_{i,q} \ge 0; \\ |A_{i,q}| \max(0, m-d_i(q)), & \text{if } A_{i,q} < 0, \end{cases}
\end{equation}
where $m$ is the predefined separation margin. Minimizing this objective encourages high-performing agents to shrink their projection distance $d_i(q)$, directly increasing their exponential familiarity score $f_i(q)$, and vice versa.

The co-evolution framework is agnostic to how agents and the router are parameterized. In practice, \sysname~supports flexible configurations: sharing a single base backbone across agents via independent LoRA adapters \cite{hu2021loralowrankadaptationlarge} for storage and memory efficiency, or deploying heterogeneous base models to exploit diverse agent capabilities.

\section{Experiments}
The empirical evaluation is designed to answer the following research questions:
\begin{itemize}
    \item \textbf{RQ1}: Does \sysname~outperform static agent routing baselines and fixed orchestration post-training paradigms?
    \item \textbf{RQ2}: How effectively does \sysname~adapt to heterogeneous base model pools?
    \item \textbf{RQ3}: Is the predictive familiarity estimator superior to direct reward estimation or multi-class classification approaches for query allocation?
    \item \textbf{RQ4}: Can the cumulative threshold adaptive routing effectively balance task performance and the number of activated agents?
    \item \textbf{RQ5}: How does \sysname~explicitly induce different capability specialization among agents?
\end{itemize}

\subsection{Experimental Setup}
\paragraph{Datasets.}
We evaluate across four domains: mathematical reasoning (GSM8k \cite{cobbe2021gsm8k}, Hendrycks MATH \cite{hendrycks2021measuring}, DAPO-MATH-17k \cite{yu2025dapoopensourcellmreinforcement}), code generation (MBPP \cite{austin2021program}, Eurus-2-Code \cite{yuan2024free}, TACO \cite{li2023taco}), instruction following (MAGPIE-IF \cite{xu2025magpie}, RLVR-IFEval \cite{lambert2024tulu}), and general reasoning (BIG-bench Hard \cite{suzgun2023challenging}). We train both the agents and the router using the training sets and report performance on their independently partitioned test sets. Out-of-distribution (OOD) generalization is evaluated on IFEval \cite{zhou2023instruction}, HumanEval \cite{chen2021evaluating}, AGIEval \cite{zhong2023agieval}, ARC-c \cite{clark2018thinksolvedquestionanswering}, LogicBench \cite{parmar2024logicbenchsystematicevaluationlogical}, and OlympiadBench \cite{he2024olympiadbench}, without any additional fine-tuning.

\paragraph{Baselines.}
We compare \sysname~against static agent orchestration optimization and predetermined-workflow agent fine-tuning baselines. Orchestration optimization baselines include \textbf{ICL-Router} \cite{wang2026icl}, which constructs capability profiles within a reconstructed latent space; \textbf{LinUCB} \cite{poon2026online}, which formulates query routing as a contextual bandit problem to dynamically select agents; and \textbf{RouteMoA} \cite{wang2026routemoa}, which employs a contrastive-learning scorer to activate multiple expert models. Agent fine-tuning baselines include the single-agent reinforcement learning baseline \textbf{GSPO} \cite{zheng2025groupsequencepolicyoptimization}; \textbf{MAPoRL} \cite{park2025maporlmultiagentpostcotrainingcollaborative}, which trains agents within a multi-agent debate framework using task and cross-agent correction rewards; and \textbf{AT-GRPO} \cite{zhao2026strongermasmultiagentreinforcementlearning}, which applies Group Relative Policy Optimization within a predefined tree-structured MoA architecture.

\paragraph{Implementation Settings.}
For the sharing base model setting, we configure a population of $N=4$ agents, equipping each agent with an independent Low-Rank Adaptation (LoRA) module. For orchestration baselines, we simulate four experts by applying role-specific prompt instructions to the base model. This architecture allows us to deploy a 4-agent MoA system on a 4B parameter base model with a highly efficient aggregate footprint of only $\sim$4.4B parameters. For the heterogeneous model setting, we configure an ensemble of $N=3$ agents by directly calling three distinct open-source base models: Qwen3-4B \cite{yang2025qwen3}, Llama-3.2-3B-Instruct \cite{grattafiori2024llama} and Phi-4-mini-Instruct \cite{abouelenin2025phi}. To maintain a fair evaluation across all baselines, we enforce identical generation hyper-parameters as well as task-specific reward functions throughout all training and inference runs as detailed in the Technical Appendix.

\subsection{Main Results (RQ1)}
\begin{table}[t]
\centering
\scriptsize
\setlength{\tabcolsep}{1.1pt} 
\begin{tabular}{l cccccc c}
\toprule
{\textbf{Method}} & 
{\fontsize{6}{6}\selectfont \textbf{IFEval}} & 
{\fontsize{6}{6}\selectfont \textbf{HumanEval}} & 
{\fontsize{6}{6}\selectfont \textbf{AGIEval}} & 
{\fontsize{6}{6}\selectfont \textbf{ARC-c}} & 
{\fontsize{6}{6}\selectfont \textbf{LogicBench}} & 
{\fontsize{6}{6}\selectfont \textbf{Olympiad}} & 
{\fontsize{6}{6}\selectfont \textbf{Avg.}} \\
\midrule
% \multicolumn{8}{l}{\textit{\textbf{Base Model: Qwen3-4B}}} \\
\textit{\textbf{Qwen3-4B}}           & \underline{79.3} & 36.4 & 64.0 & 91.7 & 17.0 & 29.7 & 53.0 \\
\hspace{0.5em}+ ICL-Router         & 78.0 & 37.0 & 67.6 & 91.1 & 17.1 & 35.5 & 54.4 \\
\hspace{0.5em}+ LinUCB             & 77.6 & 30.5 & 64.0 & 92.7 & 17.4 & 29.7 & 52.0 \\
\hspace{0.5em}+ RouteMoA           & \textbf{83.0} & 44.2 & 67.5 & \textbf{94.4} & 25.7 & 32.5 & 57.9 \\
\hspace{0.5em}+ GSPO               & 79.1 & 78.6 & 68.2 & 92.5 & 65.4 & \underline{39.2} & 70.5 \\
\hspace{0.5em}+ MAPoRL             & 78.7 & \textbf{85.1} & 66.4 & \underline{93.3} & \underline{65.5} & 34.2 & 70.5 \\
\hspace{0.5em}+ AT-GRPO       & 78.6 & \underline{84.4} & \underline{69.1} & {93.2} & 64.8 & {38.9} & \underline{71.5} \\
\rowcolor{gray!10} \textbf{\hspace{0.5em}+ \sysname} & \textbf{83.0} & {82.5} & \textbf{71.3} & {92.7} & \textbf{65.6} & \textbf{41.6} & \textbf{72.8} \\
% \addlinespace[0.5em]
\midrule
% \multicolumn{8}{l}{\textit{\textbf{Base Model: Llama-3.2-3B-Instruct}}} \\
{\fontsize{6}{6}\selectfont \textit{\textbf{Llama-3.2-3B-Instruct}}}          & 62.3 & 48.7 & 29.7 & 71.6 & 37.2 & 8.2 & 43.0 \\
\hspace{0.5em}+ ICL-Router         & 69.7 & 49.4 & 35.8 & 72.6 & 41.6 & 12.0 & 46.9 \\
\hspace{0.5em}+ LinUCB             & 67.8 & 48.1 & 36.0 & 73.6 & 41.5 & 11.6 & 46.4 \\
\hspace{0.5em}+ RouteMoA           & 69.7 & \underline{61.7} & \underline{43.4} & \underline{79.6} & \textbf{49.3} & 13.7 & \underline{52.9} \\
\hspace{0.5em}+ GSPO               & 58.0 & {55.8} & {42.3} & 78.3 & 44.8 & 15.3 & 49.1 \\
\hspace{0.5em}+ MAPoRL             & 53.0 & 51.9 & 41.8 & {79.2} & \underline{44.9} & 13.7 & 47.4 \\
\hspace{0.5em}+ AT-GRPO       & \underline{71.0} & 50.0 & 41.6 & 78.7 & \textbf{49.3} & \underline{15.6} & {51.0} \\
\rowcolor{gray!10} \textbf{\hspace{0.5em}+ \sysname} & \textbf{72.1} & \textbf{62.3} & \textbf{46.2} & \textbf{80.2} & \underline{44.9} & \textbf{17.4} & \textbf{53.9} \\
% \addlinespace[0.5em]
\midrule
% \multicolumn{8}{l}{\textit{\textbf{Base Model: Phi-4-mini-Instruct}}} \\
\textit{\textbf{Phi-4-mini-Instruct}}          & 69.9 & 57.8 & 50.5 & 71.2 & 18.0 & 24.3 & 48.6 \\
\hspace{0.5em}+ ICL-Router         & 67.1 & 66.2 & 56.9 & 71.5 & 27.6 & 27.9 & 52.9 \\
\hspace{0.5em}+ LinUCB             & 70.4 & 66.9 & 56.4 & 69.9 & 33.0 & 28.3 & 54.2 \\
\hspace{0.5em}+ RouteMoA           & 70.2 & \underline{79.9} & \underline{59.9} & 82.8 & 27.1 & \textbf{29.8} & 58.3 \\
\hspace{0.5em}+ GSPO               & 68.9 & 70.8 & 58.6 & 87.8 & \underline{35.9} & 27.6 & 58.3 \\
\hspace{0.5em}+ MAPoRL             & 60.6 & \textbf{80.5} & 58.8 & 88.1 & 34.1 & 28.3 & 58.4 \\
\hspace{0.5em}+ AT-GRPO       & \underline{73.9} & 74.0 & 59.6 & \underline{88.4} & 35.3 & \underline{29.0} & \underline{60.0} \\
\rowcolor{gray!10} \textbf{\hspace{0.5em}+ \sysname} & \textbf{74.7} & {77.3} & \textbf{60.7} & \textbf{89.6} & \textbf{42.6} & \textbf{29.8} & \textbf{62.5} \\
\midrule
% \multicolumn{11}{l}{\textit{\textbf{Base Model: Gemma-3-4B-it}}} \\
\textit{\textbf{Heterogeneous pool}}          &  &  &  &  &  &  &    \\
\hspace{0.5em}+ ICL-Router         & 74.9 & 50.0 & 60.1 & 89.3 & 18.9 & 31.0 & 54.0 \\
\hspace{0.5em}+ LinUCB            & 77.6 & 52.6 & 64.0 & 92.3 & 17.2 & 30.8 & 55.8 \\
\hspace{0.5em}+ RouteMoA          & 78.6 & 53.9 & 61.4 & 92.4 & 26.1 & 30.7 & 57.2 \\
\hspace{0.5em}+ MAPoRL            & 68.9 & \underline{78.6} & 62.3 & 89.6 & \underline{54.5} & 31.8 & 64.3 \\
\hspace{0.5em}+ AT-GRPO      & \underline{81.3} & \textbf{83.8} & \underline{67.8} & \textbf{92.9} & 50.9 & \underline{37.6} & \underline{69.1} \\
\rowcolor{gray!10} \textbf{\hspace{0.5em}+ \sysname} & \textbf{83.4} & {77.3} & \textbf{71.1} & \underline{92.7} & \textbf{62.4} & \textbf{44.1} & \textbf{71.8}\\
\bottomrule
\end{tabular}
\caption{Comparison of \sysname~with baselines on out-of-distribution (OOD) tasks.}
\label{tab:ood_results}
\end{table}
Tables~\ref{tab:main_results} and~\ref{tab:ood_results} compare \sysname~with orchestration optimization methods and fixed-workflow fine-tuning approaches for both in-distribution (ID) and out-of-distribution (OOD) tasks. \sysname~achieves the highest average performance across all three base models, demonstrating the broad applicability of our co-evolutionary framework across different model architectures and capacities.

For in-distribution tasks, our framework achieves substantial average gains. This improvement highlights the benefit of dynamically allocating training queries based on evolving agent capabilities, which enables more targeted policy updates and complementary specialization. For out-of-distribution benchmarks, \sysname~still achieves the highest average transfer scores. These results suggest that the learned capability differentiation can transfer beyond the training distribution, allowing the agents to leverage their learned competencies on unseen tasks.

\subsection{Adaptation to Heterogeneous Models (RQ2)}

We evaluate \sysname~on a heterogeneous ensemble combining Qwen3-4B, Llama-3.2-3B, and Phi-4-mini. As shown in the bottom rows of Tables~\ref{tab:main_results} and~\ref{tab:ood_results}, \sysname~still outperforms all baselines, achieving the highest average scores for both ID tasks and OOD transfer. 

Fixed-workflow paradigms often force diverse architectures into an unnatural consensus, which can dilute their unique pre-training strengths. In contrast, \sysname~leverages semantic familiarity scores to dynamically route each query to the most suitable backbone model. This capability-aware alignment is particularly effective for complex reasoning. As a result, our framework achieves substantial performance gains over the strongest post-training baselines on exceptionally challenging reasoning datasets (e.g., DAPO-MATH, LogicBench and OlympiadBench). These findings suggest that \sysname~exhibits strong adaptability across different model architectures, effectively turning model diversity into a collaborative advantage.

\begin{table}[t]
\centering
\scriptsize
\setlength{\tabcolsep}{3.8pt}
\begin{tabular}{l ccc}
\toprule
\textbf{Method / Configuration} & \textbf{ID Avg.} & \textbf{OOD Avg.} & \textbf{Avg. Tokens} \\
\midrule
\rowcolor{gray!10} \sysname~(Adaptive Threshold) & 63.2 & {72.8} & {367.77} \\
\midrule
\textit{\textbf{Routing Metric (RQ3)}} & & & \\
Direct Reward Regression & 60.8 & 70.1 &  \\
Multi-Class Classification & 61.3 & 70.4 &  \\
\midrule
\textit{\textbf{Allocation Strategy (RQ4)}} & & & \\
\sysname~(Fixed Top-1) & 61.4 & 70.1 & {325.70} \\
\sysname~(Fixed Top-2) & {63.3} & 72.5 & 666.37 \\
\bottomrule
\end{tabular}
\caption{Ablation results on Qwen3-4B.}
\label{tab:ablation_results}
\end{table}

\subsection{Ablation on Familiarity Score (RQ3)}
\begin{figure}
    \centering
    \includegraphics[width=\linewidth]{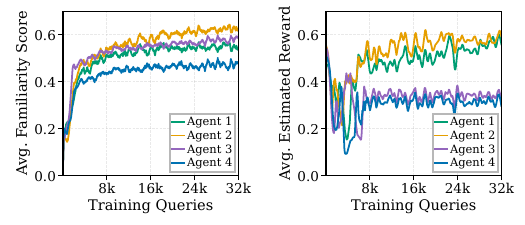}
    \caption{Comparison of routing metric dynamics.}
    \label{fig:stability_comparison}
\end{figure}
We compare our predictive familiarity estimator against two alternative routing mechanisms based on intermediate hidden states. As reported in Table~\ref{tab:ablation_results}, direct reward regression and reward-driven multi-class classification both fall noticeably short of \sysname.

Figure~\ref{fig:stability_comparison} explains this gap by plotting familiarity scores and estimated rewards across training queries. While estimated rewards exhibit oscillations, our familiarity score evolves smoothly and converges steadily. Direct reward regression and reward-driven optimization fail because rewards are inherently unstable. As agent policies evolve during training, their performance on identical prompts constantly changes. In contrast, our predictive familiarity estimator translates these dynamic shifts into a geometric push-and-pull distance between two network heads. This mechanism naturally adapts to continual learning dynamics, ensuring stable sample allocation and deeper specialization.

\subsection{Ablation on Adaptive Routing (RQ4)}
We evaluate how our cumulative-threshold adaptive routing strategy balances reasoning accuracy against computational overhead. Table~\ref{tab:ablation_results} compares our dynamic allocation mechanism against fixed Top-$k$ activation strategies on Qwen3-4B.

Within our framework, enforcing a fixed Top-2 routing strategy yields strong reasoning accuracy but incurs high token costs. Conversely, a fixed Top-1 strategy reduces computational overhead, however, it suffers a notable performance drop because a single agent struggles to resolve complex queries alone. Our cumulative-threshold mechanism resolves this trade-off. The router generally activates multiple agents only for complex problems that exceed an individual agent's expertise. Compared with fixed Top-$2$ routing, \sysname~reduces average generation tokens by approximately 45\% while retaining comparable performance.

\subsection{Capability Specialization Analysis (RQ5)}
\begin{figure}[t]
\centering
\begin{minipage}{0.48\linewidth}
    \centering
    \includegraphics[width=\linewidth]{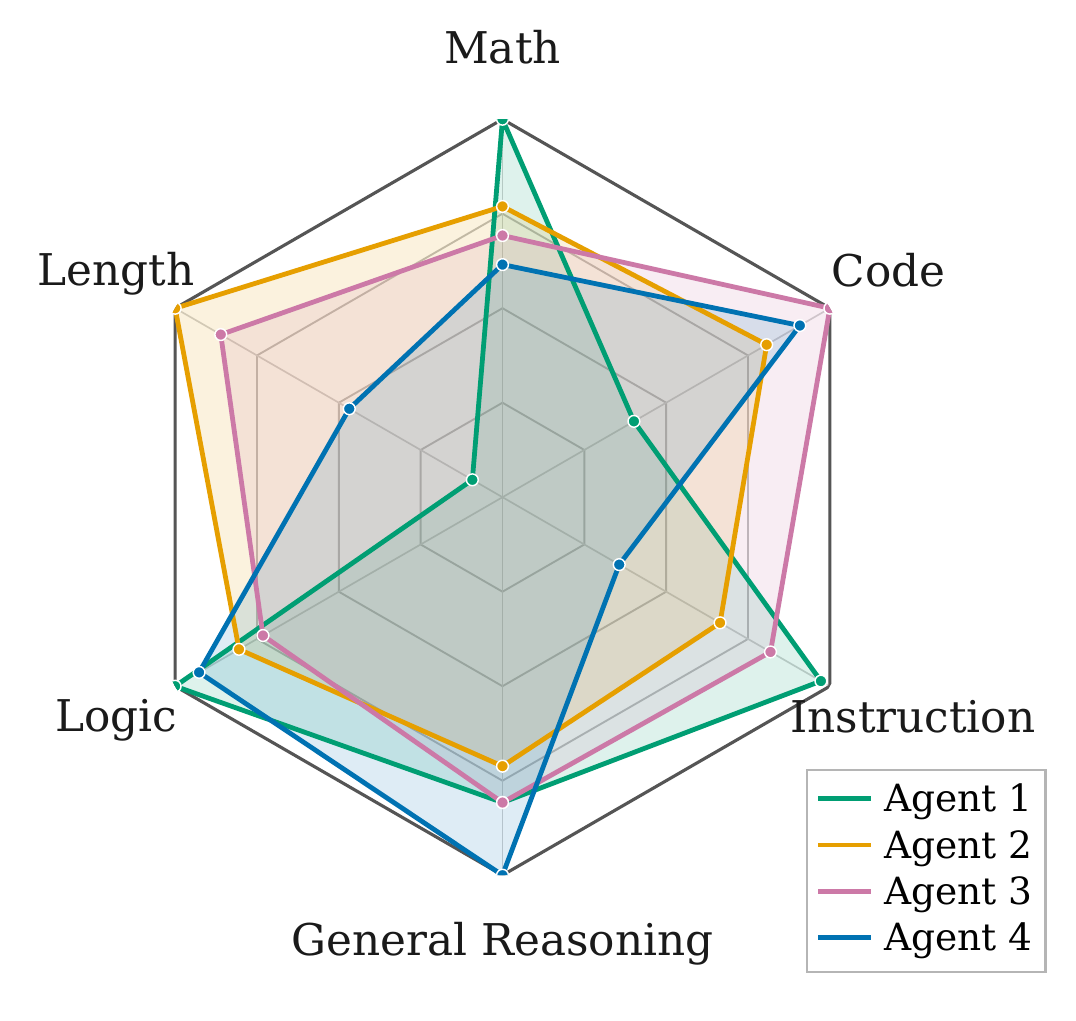}
\end{minipage}
\hfill
\begin{minipage}{0.48\linewidth}
    \centering
    \includegraphics[width=\linewidth]{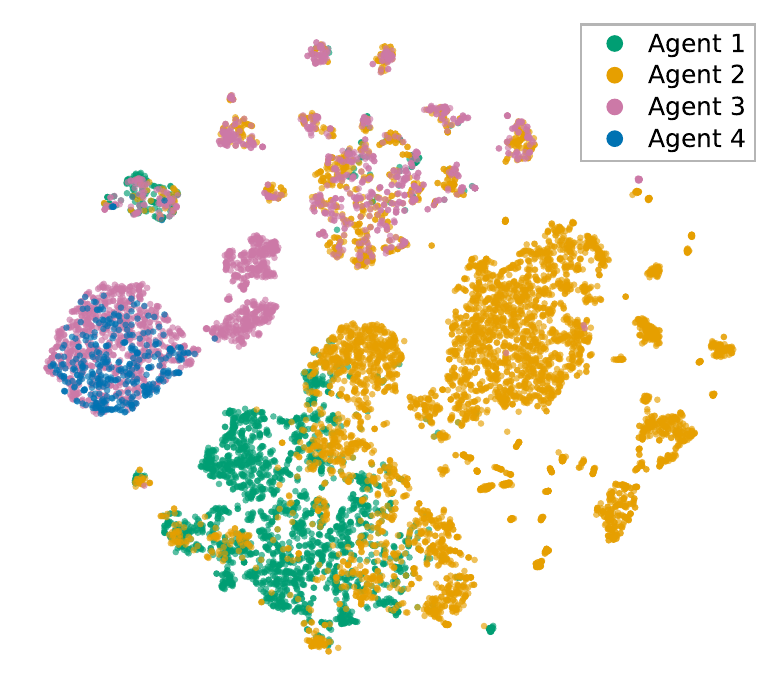}
\end{minipage}
\caption{Capability analysis of Qwen3-4B agents.}
\label{fig:specialization}
\end{figure}
We investigate how our closed-loop co-evolution explicitly promotes distinct problem-solving specialization without human intervention or role assignment. We analyze the emergent behavioral profiles of the four Qwen3-4B agents.

As shown in Figure~\ref{fig:specialization} (Left), the agent population spontaneously differentiated into complementary specialists. Due to long chain-of-thought reasoning, Agent 2 was allocated the majority of queries as a primary solver. To supplement this foundational reasoning capacity, Agent 1 mastered concise math and logic, Agent 3 focused on coding, and Agent 4 specialized in general reasoning. Figure~\ref{fig:specialization} (Right) explains this emergence through a t-SNE visualization of intermediate hidden states. Queries routed to the same agent form tightly clustered semantic neighborhoods. By consistently allocating similar prompts to the most receptive model during training, our predictive familiarity estimator drives deep policy differentiation.

\section{Conclusion}
This paper introduces \sysname, which co-evolves query routing with continually learning LLM agents. By evaluating agent competence via familiarity scores based on hidden-state semantics and applying cumulative-threshold routing, \sysname~balances efficiency with reasoning accuracy while inducing domain specialization, significantly outperforming static routing and fixed-workflow training baselines. 

For future work, extending our single-turn semantic routing to multi-turn interactions or multi-agent debate could better support iterative problem-solving. Additionally, upgrading our familiarity-weighted voting to advanced answer aggregation mechanisms, such as an adaptive meta-thinker, promises to further enhance collective reasoning.

\bibliography{refs} 

\appendix
\section{Algorithm Pseudocode \& Architecture Details}
\label{app:algorithm}

\subsection{Co-Evolutionary Training Procedure}
\label{app:training-procedure}
\begin{algorithm}[h]

\caption{\sysname~Co-Evolutionary Training}

\label{alg:cera-moa}

\begin{algorithmic}[1]

\REQUIRE Training dataset $\mathcal{D}$; agent policies $\{\pi_{\theta_i}\}_{i=1}^{N}$; frozen target heads $\{\bar g_i\}_{i=1}^{N}$; trainable predictor heads $\{g_i\}_{i=1}^{N}$; cumulative threshold $\tau$; separation margin $m$

\STATE Initialize an empty LIFO sample buffer $\mathcal{B}_i$ for each agent $i \in \{1, \dots, N\}$

\FOR{each orchestration step}

    \STATE Sample a query batch $\mathcal{Q}\subset\mathcal{D}$ and extract frozen semantic representations $\{h(q):q\in\mathcal{Q}\}$

    \FOR{each query $q\in\mathcal{Q}$}

        \STATE Compute raw familiarity scores $\{f_i(q)\}_{i=1}^{N}$ via Eq.~\ref{eq:appendix-familiarity}

        \STATE Calculate exploration routing scores $\{s_i(q)\}_{i=1}^{N}$ via Eq.~\ref{eq:appendix-exploration}

        \STATE Sort agents in descending order by $s_i(q)$

        \STATE Activate minimal prefix subset $S_q$ satisfying $\sum_{i\in S_q}f_i(q)\geq\tau$ (or activate all $N$ agents if infeasible)

        \STATE Push tuple $(q, h(q))$ into LIFO buffer $\mathcal{B}_i$ for every activated agent $i\in S_q$

    \ENDFOR

    \FOR{each agent $i$ with sufficient samples in $\mathcal{B}_i$}

        \STATE Pop a micro-batch $Q_i$ from LIFO buffer $\mathcal{B}_i$

        \STATE Load adapter weights $\theta_i$ and sample $G$ trajectory rollouts per query $q\in Q_i$

        \STATE Evaluate task rewards $\{R(q,a_{i,q}^{(g)})\}_{g=1}^{G}$ and update policy $\theta_i$ via joint DAPO--GSPO optimization

        \STATE Record empirical mean reward $\bar r_{i,q}=\frac{1}{G}\sum_{g=1}^{G}R(q,a_{i,q}^{(g)})$ for synchronous router supervision

    \ENDFOR

    \FOR{each agent $i$ participating in routed queries}

        \STATE Compute relative advantage $A_{i,q}=\bar r_{i,q}-\frac{1}{|S_q|}\sum_{j\in S_q}\bar r_{j,q}$ (if $|S_q|=1$, set $A_{i,q}=\bar r_{i,q}$)

        \STATE Update predictor head $g_i$ via Eq.~\ref{eq:appendix-router-loss}, keeping $\bar g_i$ and the backbone model frozen

    \ENDFOR

\ENDFOR

\end{algorithmic}

\end{algorithm}

We formally detail the end-to-end training methodology of \sysname~in Algorithm~\ref{alg:cera-moa}. Our co-evolutionary framework is inherently agnostic to the underlying parameterization of the agent population $\left\{\pi_{\theta_i}\right\}_{i=1}^{N}$, supporting two primary deployment paradigms. In the \textit{homogeneous setting}, agents are instantiated as independent, lightweight Low-Rank Adaptation (LoRA) modules over a single shared backbone, achieving high parameter efficiency and low memory overhead during multi-agent rollouts. In the \textit{heterogeneous setting}, each agent is powered by a distinct base model, allowing the router to actively exploit architectural diversity and complementary pre-training competencies. Crucially, query features are extracted from the frozen router backbone by concatenating the final non-padding hidden states from intermediate layers. Throughout the closed-loop optimization, only the agents' policy parameters $\theta_i$ and the router's lightweight predictor heads $g_i$ are updated.

\paragraph{Feature Extraction and Familiarity Estimation.}
For an incoming query $q$, the router tokenizes the prompt formatted for its specific task class and extracts the final non-padding hidden state from intermediate layers. Following our core findings, we concatenate the hidden states from layers $\lfloor L/2 \rfloor$ and $\lfloor 3L/4 \rfloor$, yielding a rich semantic representation $h(q)\in\mathbb{R}^{2H}$ for a backbone with hidden dimension $H$ and total depth $L$. This mid-layer extraction captures deep syntactic and reasoning structures without overfitting to immediate next-token prediction dynamics.

To evaluate agent-query compatibility, each agent $i$ is assigned a trainable predictor head $g_i$ and a permanently frozen target head $\bar g_i$, both using an identical Multi-Layer Perceptron (MLP) architecture:
\begin{equation}
    \operatorname{LN}(h)\rightarrow\operatorname{Linear}(2H, 512)
\rightarrow\operatorname{SiLU}\rightarrow\operatorname{Linear}(512, 256).
\end{equation}
To establish a stable and diverse reference space across the population, the target heads $\left\{\bar g_i\right\}_{i=1}^{N}$ are initialized orthogonally and remain strictly frozen. By projecting the representations onto a unit hypersphere, the semantic projection distance $d_i(q)$ and the resulting familiarity score $f_i(q)$ are formulated as:
\begin{equation}
\begin{aligned}
d_i(q) &=
\left\|
\frac{g_i(h(q))}{\left\|g_i(h(q))\right\|_2}
-\frac{\bar g_i(h(q))}{\left\|\bar g_i(h(q))\right\|_2}
\right\|_2,\\
f_i(q) &= \exp\left(-\lambda d_i(q)\right),
\end{aligned}
\label{eq:appendix-familiarity}
\end{equation}
where $\lambda>0$ is the temperature hyperparameter regulating distance sensitivity. Consequently, a smaller projection distance maps exponentially to a higher familiarity score, signaling stronger compatibility alignment.

\paragraph{Exploration-Augmented Query Routing.}
During co-evolutionary training, query allocation must actively explore under-utilized agents to prevent policy starvation and encourage diversity, without compromising the actual competence required to solve the task. To balance these objectives, we make a strict distinction between agent \textit{prioritization} and the subset \textit{activation threshold}. To foster exploration, agents are ranked by an exploration-augmented routing score $s_i(q)$:
\begin{equation}
s_i(q)=f_i(q)+
c_{\mathrm{ucb}}\sqrt{\frac{\log(T)}{n_i+1}}+
w_{\mathrm{ent}}\bar H_i,
\label{eq:appendix-exploration}
\end{equation}
where $T$ denotes the total number of routed queries, $n_i$ represents the historical training sample count allocated to agent $i$, and $\bar H_i$ is its running mean generation entropy. While the ranking order is dictated by $s_i(q)$, the activation criterion $\tau$ is evaluated strictly against the raw familiarity scores. Specifically, the router selects the minimal prefix subset $S_q$ ordered by $s_i(q)$ such that:
\begin{equation}
    \sum_{i\in S_q}f_i(q)\geq\tau.
\end{equation}
If the entire population fails to reach $\tau$, all agents are activated. This critical design prevents exploration bonuses from inflating perceived competence coverage.

\paragraph{Agent Policy Optimization.}
Capability differentiation is driven by each agent independently optimizing its reasoning policy using training samples dynamically allocated by the router. When an agent $i$ accumulates a sufficient micro-batch $Q_i$ in its LIFO buffer $\mathcal{B}_i$, it samples $G$ distinct trajectory rollouts $\left\{a_{i,q}^{(1)}, \dots, a_{i,q}^{(G)}\right\}$ for each query $q \in Q_i$. Let $R\left(q, a_{i,q}^{(g)}\right)$ denote the scalar reward evaluated by the task-specific verifier. 

To stabilize multi-sample reinforcement learning and mitigate reward scale variance across heterogeneous tasks, we employ Dynamic Sampling Policy Optimization (DAPO). Specifically, we compute the empirical mean $\mu_{i,q}$ and standard deviation $\sigma_{i,q}$ over the $G$ sampled responses for query $q$. The intra-agent normalized advantage $\hat{A}_{i,q}^{(g)}$ for the $g$-th trajectory is formulated as:
\begin{equation}
\hat{A}_{i,q}^{(g)} = \frac{R\left(q, a_{i,q}^{(g)}\right) - \mu_{i,q}}{\sigma_{i,q} + 10^{-4}}.
\label{eq:appendix-agent-adv}
\end{equation}
Integrating Group Sequence Policy Optimization (GSPO) importance sampling ratios effectively optimizes long-horizon chain-of-thought reasoning while mitigating token-level variance. For a generated completion $a_{i,q}^{(g)}=\left(y_{1},...,y_{S}\right)$ of length $S$, we define the sequence-level importance sampling ratio $\rho_{i,q}^{(g)}$ averaged over all valid generation tokens:
\begin{equation}
\rho_{i,q}^{(g)} = \exp \left( \frac{1}{S} \sum_{t=1}^{S} \log \frac{\pi_{\theta_i}\left(y_t \mid q, y_{<t}\right)}{\pi_{\theta_i^{\mathrm{old}}}\left(y_t \mid q, y_{<t}\right)} \right).
\end{equation}
We then define the clipped surrogate objective $s_{i,q}^{(g)}$ for the $g$-th completion as:
\begin{equation}
s_{i,q}^{(g)} = -\min \left( \rho_{i,q}^{(g)} \hat{A}_{i,q}^{(g)}, \, \operatorname{clip}\left(\rho_{i,q}^{(g)}, 1-\epsilon, 1+\epsilon\right) \hat{A}_{i,q}^{(g)} \right),
\end{equation}
where $\epsilon$ is the clipping threshold. To prevent policy degeneration and reward hacking, we apply a token-level Kullback–Leibler (KL) divergence penalty against a fixed reference policy $\pi_{\mathrm{ref}}$. Defining the log-ratio at token step $t$ as:
\begin{equation}
x_t = \log \pi_{\mathrm{ref}}(y_t \mid q, y_{<t}) - \log \pi_{\theta_i}(y_t \mid q, y_{<t}),
\end{equation}
the token-level KL penalty utilizes the estimator:
\begin{equation}
D_{\mathrm{KL}, t} = \exp(x_t) - x_t - 1.
\end{equation}
Finally, the objective for agent $i$ is normalized by the total number of active (non-padded) completion tokens across the accumulated micro-batch:
\begin{equation}
\mathcal{L}_{\mathrm{agent}}(\theta_i) = \frac{\sum_{q \in Q_i} \sum_{g=1}^{G} \sum_{t=1}^{S_g} m_{q,g,t} \left( s_{i,q}^{(g)} + \beta D_{\mathrm{KL}, t} \right)}{\sum_{q \in Q_i} \sum_{g=1}^{G} \sum_{t=1}^{S_g} m_{q,g,t}},
\label{eq:appendix-agent-loss}
\end{equation}
where $m_{q,g,t} \in \{0, 1\}$ masks out padding tokens, and $\beta$ serves as the regularization scaling hyperparameter.

\paragraph{Router Optimization.}
As individual agent policies $\left\{\pi_{\theta_i}\right\}_{i=1}^{N}$ evolve, their semantic competence boundaries shift. To maintain synchronization, the router is continually optimized using the feedback of the participating agents. Unlike the intra-agent advantage $\hat{A}_{i,q}^{(g)}$ used for policy updates, the router evaluates a \textit{cross-agent relative advantage} $A_{i,q}$ that measures how agent $i$'s average performance compares against the collective consensus of the active subset $S_q$:
\begin{equation}
A_{i,q} = \bar r_{i,q} - \frac{1}{|S_q|} \sum_{j \in S_q} \bar r_{j,q},
\end{equation}
where $\bar r_{i,q} = \frac{1}{G} \sum\limits_{g=1}^{G} R\left(q, a_{i,q}^{(g)}\right)$ is the empirical mean reward of agent $i$. For solo-activated agents ($\left|S_q\right|=1$), we set $A_{i,q} = \bar r_{i,q}$ to reward standalone mastery. 

Geometrically, the router objective transforms these relative reward signals into a conditional push-and-pull optimization within the fixed reference space:
\begin{equation}
\mathcal{L}_{\mathrm{router}}=
\sum_{i,q}
\begin{cases}
A_{i,q}d_i(q), & \text{if } A_{i,q}\geq0,\\
|A_{i,q}|\max(0, m-d_i(q)), & \text{if } A_{i,q}<0,
\end{cases}
\label{eq:appendix-router-loss}
\end{equation}
where $m>0$ defines the separation margin. Minimizing Equation~(\ref{eq:appendix-router-loss}) pulls the predictor heads of outperforming agents ($A_{i,q} \ge 0$) closer to their fixed anchors, thereby driving up their future familiarity scores for similar queries. Conversely, for underperforming agents ($A_{i,q} < 0$), the hinge mechanism pushes the predictor heads away from their anchors, but strictly bounds this repulsion to a maximum projection distance of $m$. This bounded separation is critical to prevent familiarity score collapse: without the margin $m$, unbounded push-away optimization on negative advantages would induce excessively large gradients, driving the projection distance to extremes and permanently collapsing the exponential familiarity score to zero. By capping the repulsion at distance $m$, our objective dynamically suppresses the activation probability for semantically similar queries while safeguarding the numerical and geometric stability of the router's representation space. In practice, the value of $m$ is set such that when the projection distances of all agents reach this boundary, their cumulative familiarity score falls just below the required activation threshold $\tau$ (i.e., $\sum_{i=1}^{N} \exp(-\lambda m) \leq\tau$). This ensures that for universally unfamiliar or exceptionally difficult queries where no agent demonstrates competence, the cumulative threshold condition naturally fails, gracefully triggering the fallback strategy to activate the entire agent population.

\subsection{Inference and Consensus Aggregation}
\label{app:inference}

During test-time inference, all exploration dynamics are terminated ($c_{\mathrm{ucb}} = w_{\mathrm{ent}} = 0$). The router strictly applies the cumulative-threshold mechanism over learned familiarity scores $\left\{f_i(q)\right\}_{i=1}^{N}$, activating only the minimal subset of specialist agents required to satisfy $\tau$. Each selected agent generates a single independent completion. 

For final output synthesis, \sysname~deploys a domain-aware aggregation protocol. For tasks with verifiable, deterministic solutions (e.g., mathematical reasoning or choice questions), the aggregator executes \textit{familiarity-weighted majority voting}: candidate completions are grouped by their extracted final answers, and the winning consensus group is identified by summing the familiarity scores $f_i(q)$ of its contributing agents. The system then outputs the exact generation trajectory from the highest-familiarity agent within that winning group. For open-ended generation tasks where exact consensus is inapplicable, the aggregator directly outputs the response from the most familiar activated agent ($\arg\max_{i \in S_q} f_i(q)$), maximizing domain alignment with minimal computational overhead.

\section{Detailed Experimental Setup \& Reproducibility}
\label{app:reproducibility}

\subsection{Hardware and Software Infrastructure}
\label{app:infra}

All computational experiments are executed on a standardized high-performance computing cluster operating under the {Ubuntu 22.04.5 LTS} Linux distribution. The node is equipped with four NVIDIA H200 Tensor Core GPUs (each featuring 141\,GB of HBM3e memory), dual Intel Xeon Platinum 8558 processors (yielding 96 physical CPU cores), and approximately 500\,GB of host system memory. All neural network optimization and inference pipelines are implemented in Python under a CUDA-enabled PyTorch environment, utilizing mixed \texttt{bfloat16} precision to balance computational throughput and numerical stability.
\begin{table}[h]
\centering
\small
\begin{tabular}{l c}
\toprule
\textbf{Library} & \textbf{Version / Role Description} \\
\midrule
Python & 3.12 \\
PyTorch & 2.9.1 \\
Transformers & 4.57.6 \\
PEFT & 0.11.1 \\
vLLM & 0.16.0 \\
Accelerate & 1.13.0 \\
Datasets & 4.8.4  \\
Safetensors & 0.7.0  \\
SymPy & 1.14.0  \\
NumPy & 2.2.6  \\
Pandas & 2.2.2  \\
\bottomrule
\end{tabular}
\caption{Major software libraries and dependencies imported by our training, reward evaluation, and verification frameworks. TRL components are sourced from our customized repository-local implementation to support group-normalized DAPO and GSPO sequence-level importance sampling.}
\label{tab:appendix-software}
\end{table}
Scaling multi-agent reinforcement learning within hardware memory budgets relies on implementing re-entrant gradient checkpointing during policy backward passes, substantially reducing activation memory consumption. Furthermore, we integrate memory-isolated, co-located vLLM engines to accelerate trajectory rollout generation during training. In the homogeneous setting, four independent LoRA adapter modules are co-hosted over a single shared LLM backbone, sharing fixed base model weights in GPU memory. In the heterogeneous setting, isolating each distinct base model-adapter pair within a dedicated worker process prevents CUDA context contention and memory fragmentation between the local training stack and the vLLM inference engine.

Regarding the training computational expenditure, advancing the co-evolutionary training by 1,000 optimization steps, which corresponds to processing 16,000 training queries and sampling $G=8$ trajectory rollouts per query, requires approximately one H200 GPU day. 

Table~\ref{tab:appendix-software} outlines the primary software libraries and framework dependencies required to execute our training, verification, and data-loading pipelines.

\subsection{Hyper-Parameter Configuration}
\label{app:hparams}

\begin{table*}[h]
\centering
\small
\begin{tabular}{l p{1.3\columnwidth}}
\toprule
\textbf{Hyper-Parameter Category} & \textbf{Configured Value \& Description} \\
\midrule
Random seed &42\\
Agent population size ($N$) & $N=4$ for shared-backbone settings; $N=3$ for heterogeneous setting \\
LoRA hyper-parameters & Rank $r=16$, Alpha $\alpha=32$, Dropout probability $0.05$ \\
LoRA target modules & \texttt{q\_proj, k\_proj, v\_proj, o\_proj, gate\_proj, up\_proj, down\_proj} \\
Learning rates & Router predictor head learning rate: $1 \times 10^{-4}$; Agent adapter learning rate: $3 \times 10^{-5}$ \\
Agent Learning-rate schedule & Cosine decay schedule with 600 warm-up steps; Minimum learning rate floor: $2 \times 10^{-5}$ \\
Router routing parameters & Temperature $\lambda=2.0$, Threshold $\tau=0.7$, Separation margin $m=1.0$ \\
Exploration coefficients & UCB bound $c_{\mathrm{ucb}}=2.0$; Entropy weight $w_{\mathrm{ent}}=2.0$ (shared) / $0.2$ (heterogeneous) \\
TRL optimization objective & DAPO loss with group-normalized rewards and GSPO importance sampling ratios \\
Trajectory rollout count ($G$) & $G=8$ independent completion rollouts per allocated training prompt \\
Regularization & KL divergence penalty weight $\beta=1 \times 10^{-4}$; Policy ratio clip margin $\epsilon=0.2$ \\
Batching \& accumulation & Per-device batch size: 16 prompts; Gradient accumulation steps: $A=8$ \\
Training duration \& limits & Maximum optimizer updates: 6,000 steps; Trainer steps per rollout generation: 2 \\
Optimizer specifics & AdamW optimizer with $\beta_1=0.9, \beta_2=0.95$; Agent weight decay: $1 \times 10^{-3}$ \\
Gradient clipping & Router predictor head gradient norm clipped at $1.0$; Agent gradient norm clipped at $1.0$ \\
Generation parameters & Sampling temperature: $1.0$; Top-$p$ (nucleus): $0.95$; Top-$k$: $50$; Repetition penalty: $1.0$ \\
Sequence length limits & Maximum input prompt: 800 tokens; Maximum generated completion: 1,000 tokens \\
\bottomrule
\end{tabular}
\caption{Primary hyper-parameter configurations for CERA-MoA training and inference.}
\label{tab:appendix-hparams}
\end{table*}
Table~\ref{tab:appendix-hparams} provides the comprehensive hyper-parameter schedule. Crucially, our co-evolutionary system decouples query throughput from fixed agent epoch cycles. In each orchestration round, the router samples a global query batch of size $B_{\mathrm{global}} = \frac{B_{\mathrm{device}} \times A}{G} = \frac{16 \times 8}{8} = 16$ distinct prompts, where $B_{\mathrm{device}}$ denotes the per-device batch size, $A$ is the gradient accumulation factor, and $G$ represents the number of sampled completion trajectories per prompt. An individual agent policy $\theta_i$ only executes a DAPO--GSPO optimization step when its local LIFO sample buffer $\mathcal{B}_i$ accumulates a complete micro-batch ($\vert{}Q_i\vert{} = 16$). Consequently, the training update frequency per agent is inherently \textit{adaptive}: agents exhibiting higher familiarity and competence in specific semantic domains receive richer training allocation, naturally accelerating their capability specialization without requiring manual curriculum schedules.

\subsection{Prompt Design and Zero-Shot Task Instructions}
\label{app:prompts}
We enforce a zero-shot, role-agnostic prompt design, rigorously evaluating whether CERA-MoA can induce genuine, data-driven domain specialization. Unlike conventional multi-agent orchestration baselines that rely heavily on prompt engineering---assigning explicit personas (e.g., "You are an expert mathematician...") or structured verification workflows---our system inputs contain only the raw user query. No overarching system prompts, role descriptors, or behavioral hints are injected into the context window. This design guarantees that the emergent capability differentiation observed in our empirical analysis (Section 4.6) is strictly driven by our predictive familiarity routing and closed-loop reinforcement learning, rather than human-engineered role bias.

To prevent uncontrolled token inflation and maintain standardized evaluation across diverse base models, any proprietary or built-in reasoning modes (e.g., native internal "thinking" or hidden chain-of-thought routines) are disabled during generation; models rely entirely on standard autoregressive step-by-step reasoning. Prompts exceeding the configured input limit (800 tokens) are deterministically truncated from the left prior to routing and execution. Figure~\ref{fig:appendix-prompts} presents the exact task-specific user formatting templates applied across all benchmark domains.
% Requires: \usepackage[most]{tcolorbox}

\begin{figure*}[h]
\centering
\small
\begin{tcolorbox}[
    colback=gray!4,
    colframe=black!70,
    arc=3pt,
    boxrule=0.8pt,
    left=8pt, right=8pt, top=8pt, bottom=8pt,
    fonttitle=\bfseries,
    coltitle=white,
    title={Task-Specific Zero-Shot Prompt Templates across Benchmark Domains}
]

{\bfseries\color{black!80}[Domain: Mathematical Reasoning]}
\par\smallskip\hrule\smallskip
\noindent\textbf{Problem:}\\
\texttt{\color{blue!70!black}\{question\}}

\smallskip
\noindent\textbf{Please reason step by step.}

\smallskip
\noindent Present the final answer in the following format:\\
\textbf{Answer:} \texttt{\textbackslash boxed\{XX\}}

\medskip

{\bfseries\color{black!80}[Domain: General Reasoning]}
\par\smallskip\hrule\smallskip
\noindent\textbf{Problem:}\\
\texttt{\color{blue!70!black}\{question\}}

\smallskip
\noindent\textbf{Please reason step by step.}

\smallskip
\noindent Present the final answer in the following format:\\
\textbf{Answer:} \texttt{\textbackslash boxed\{XX\}}

\medskip

\begin{minipage}[h]{0.48\textwidth}
{\bfseries\color{black!80}[Domain: Code Generation for a function]}
\par\smallskip\hrule\smallskip
\noindent\textbf{Problem:}\\
\texttt{\color{blue!70!black}\{question\}}

\smallskip
\noindent\textbf{Please reason step by step.}

\smallskip
\noindent Write a Python function named
\texttt{\color{blue!70!black}\{fn\_name\}} to solve the problem.

\smallskip
\noindent Present the code in:\\
\texttt{```python}\\
\texttt{Your code}\\
\texttt{```}
\end{minipage}\hfill
\begin{minipage}[h]{0.48\textwidth}
{\bfseries\color{black!80}[Domain: Code Generation]}
\par\smallskip\hrule\smallskip
\noindent\textbf{Problem:}\\
\texttt{\color{blue!70!black}\{question\}}

\smallskip
\noindent\textbf{Please reason step by step.}

\smallskip
\noindent Write Python code to solve the problem.

\smallskip
\noindent Present the code in:\\
\texttt{```python}\\
\texttt{Your code}\\
\texttt{```}
\end{minipage}

\medskip

{\bfseries\color{black!80}[Domain: Open-Ended Instruction Following]}
\par\smallskip\hrule\smallskip
\noindent\texttt{\color{blue!70!black}\{question\}}

\end{tcolorbox}
\caption{Zero-shot prompt templates used for different task families. The two code-generation blocks
correspond to function-call and standard-input/output task interfaces,
respectively. Variables \texttt{\{question\}} and \texttt{\{fn\_name\}} are
instantiated from each dataset item. Instruction-following prompts are passed
through unchanged so that their verifier-defined constraints are preserved.}
\label{fig:appendix-prompts}
\end{figure*}

\section{Task-Specific Reward Formulation}
\label{app:rewards}

Each rollout receives a task reward $r_{\mathrm{base}}\in[0,1]$ followed by a length penalty.  Let $\ell$ denote the number of generated tokens, $L_{\max}$ the maximum completion length, $\gamma\in(0,1)$ the fraction of $L_{\max}$ at which penalization begins, and $\eta\geq0$ the maximum penalty magnitude.  The final reward is
\begin{equation}
R=r_{\mathrm{base}}-
\eta\cdot
\begin{cases}
0, & \ell\leq\gamma L_{\max},\\
\dfrac{\ell-\gamma L_{\max}}{(1-\gamma)L_{\max}}, & \gamma L_{\max}<\ell<L_{\max},\\
1, & \ell\geq L_{\max}.
\end{cases}
\label{eq:appendix-reward}
\end{equation}
Thus, the penalty is zero up to $\gamma L_{\max}$ and grows linearly thereafter, reaching $\eta$ at the completion limit.  Accuracy reported at evaluation is only computed from whether $r_{\mathrm{base}}=1$.

\paragraph{Mathematical reasoning.}
For math problems, the evaluator searches for all occurrences of \texttt{\textbackslash boxed\{\}} and returns the last complete balanced-brace expression, which avoids treating an intermediate derivation as the final answer.  It then normalizes common presentation variants before symbolic comparison: surrounding delimiters and whitespace, units, thousands separators, reducible fractions, percentage markers, equivalent matrix delimiters, and multiple-choice notation.  The verifier additionally handles set-style answers, $\pm$ expansions, ratios, and mathematical expressions represented in \LaTeX.  It compares the normalized prediction and reference using exact checks followed by SymPy simplification and equivalent algebraic transformations; the process is isolated with a timeout to prevent malformed expressions from blocking training.  The binary reward is
\[
r_{\mathrm{base}}^{\mathrm{math}}=
\mathbb{I}\{\operatorname{MathEqual}(\operatorname{Extract}(a),y^\star)\}.
\]
This procedure accepts mathematically equivalent expressions while retaining exact matching when symbolic parsing is unavailable.

\paragraph{Code generation.}
The evaluator extracts the last fenced Python block when present, otherwise the raw completion.  Code is executed in an isolated subprocess under a reliability guard and a 10-second wall-clock limit covering compilation and all tests.  Both stdin/stdout programs and function-call tasks are supported.  With the enabled stepwise reward, the score is the pass fraction over the task's tests:
\[
r_{\mathrm{base}}^{\mathrm{code}}=
\frac{1}{|\mathcal{T}|}\sum_{(x,y)\in\mathcal{T}}
\mathbb{I}\{\operatorname{Run}(a,x)=y\}.
\]
Compilation failures, runtime failures, timeouts, malformed outputs, and missing code receive zero reward.  The implementation can instead use strict all-tests-pass reward.

\paragraph{Instruction following.}
Each IF item contains an instruction identifier list and its instantiated argument dictionaries.  The official IFEval \cite{zhou2023instruction} instruction registry reconstructs each verifier.  With stepwise instruction reward enabled, the reward is the fraction of satisfied constraints:
\[
r_{\mathrm{base}}^{\mathrm{IF}}=
\frac{1}{K}\sum_{k=1}^{K}
\mathbb{I}\{\operatorname{Check}_{k}(a;q,\omega_k)=1\}.
\]
Examples include exact endings, required keywords, case restrictions, sentence or bullet counts, and markup constraints.  If a registry entry is unavailable or the response is empty, the reward is zero.  The strict alternative awards one only if every constraint is satisfied.

\paragraph{General reasoning.}
General reasoning items use the same final-answer extractor as math problems, followed by task-agnostic normalization: optional ``Answer:'' prefixes and enclosing parentheses are removed, one-letter choices are uppercased, and remaining whitespace is canonicalized and lowercased.  The reward is exact match after normalization:
\[
r_{\mathrm{base}}^{\mathrm{general}}=
\mathbb{I}\{\operatorname{Normalize}(\operatorname{Extract}(a))=
\operatorname{Normalize}(y^\star)\}.
\]

\section{Dataset Specifications}
\label{app:datasets}

\subsection{Source Datasets}

\paragraph{GSM8K.}
GSM8K~\cite{cobbe2021gsm8k} contains 7,473 training and 1,319 test grade-school mathematical word problems.  Each item requires multi-step arithmetic reasoning from a natural-language question to a short numerical answer.

\paragraph{MATH.}
MATH~\cite{hendrycks2021measuring} has 7,500 training and 5,000 test competition-style problems spanning algebra, geometry, number theory, counting, and probability.  Reference answers are represented in mathematical notation and require equivalence-aware rather than string-based verification.

\paragraph{DAPO-MATH-17K.}
DAPO-MATH-17K~\cite{yu2025dapoopensourcellmreinforcement} is a verifiable mathematical reasoning collection.  The processed version contains 13,616 training items and a 500-item test split, covering a broad range of symbolic and quantitative problem types with automatically checkable answers.

\paragraph{MBPP.}
The sanitized MBPP split~\cite{austin2021program} used in our pipeline contains 163 training and 257 test Python programming tasks.  Problems specify a function-level intent and executable input--output examples, supporting direct functional correctness checks.

\paragraph{Eurus-2-Code.}
The processed Eurus-2-Code split~\cite{yuan2024free} contains 25,278 training and 1,024 test code-generation tasks.  Each task is associated with executable input--output tests, extending the code domain beyond short function synthesis.

\paragraph{TACO.}
TACO~\cite{li2023taco} contains 24,702 training and 1,000 test programming problems.  In addition to function-call tasks, it includes standard-input/standard-output problems that evaluate complete programs against executable test cases.

\paragraph{MAGPIE-IF.}
The IFEval-like instruction-following source derived from MAGPIE~\cite{xu2025magpie} provides 551,465 training prompts and a 500-item test split.  Its prompts encode constraints such as required content, output structure, length, casing, and ordering in a form that can be programmatically verified.

\paragraph{RLVR-IFEval.}
RLVR-IFEval~\cite{lambert2024tulu} is a rule-verifiable instruction-following source based on the IFEval protocol, with 14,540 training and 500 test examples.  Each instance associates a natural-language request with the instruction identifiers and arguments needed to reconstruct its verifiers.

\paragraph{BIG-bench Hard.}
BIG-bench Hard (BBH)~\cite{suzgun2023challenging} contains 6,011 training and 500 test instances in the processed split.  It covers challenging general reasoning, including symbolic manipulation, logical deduction, temporal and spatial reasoning, language understanding, and multiple-choice tasks.

\paragraph{IFEval.}
IFEval~\cite{zhou2023instruction} is an instruction-following benchmark with 541 evaluation prompts and no training split.  It measures compliance with individually and jointly specified constraints using deterministic verifiers rather than model-based judging.

\paragraph{HumanEval.}
The processed HumanEval benchmark~\cite{chen2021evaluating} contains 154 test-only Python programming problems.  Each problem specifies a function signature, a natural-language docstring, and unit tests designed to assess functional correctness.

\paragraph{AGIEval.}
The AGIEval evaluation file~\cite{zhong2023agieval} contains 1,000 test-only multiple-choice questions.  It assesses general knowledge and reasoning across standardized examinations and does not contribute training instances to our method.

\paragraph{ARC-Challenge.}
ARC-Challenge~\cite{clark2018thinksolvedquestionanswering} contains 1,172 test-only science questions in the processed evaluation split.  Questions are multiple choice and target grade-school scientific reasoning that typically requires combining facts or applying basic principles.

\paragraph{LogicBench.}
The LogicBench evaluation file~\cite{parmar2024logicbenchsystematicevaluationlogical} contains 1,520 test-only instances.  It focuses on formal and natural-language logical reasoning, including entailment, deduction, and structured constraint solving.

\paragraph{OlympiadBench.}
The OlympiadBench evaluation file~\cite{he2024olympiadbench} contains 707 test-only problems.  It evaluates high-difficulty mathematical and scientific reasoning with questions drawn from olympiad-style settings.

\subsection{Construction of Training and In-Distribution Test Sets}

The resulting training corpus contains 32,000 examples across four task families, deterministically sampled using a fixed random seed (\texttt{seed = 42}) to ensure full reproducibility: 11,200 mathematical reasoning examples (MATH: 5,000, DAPO-MATH-17k: 5,000, and GSM8K: 1,200), 9,600 code-generation examples (MBPP: 163, Eurus-2-Code: 7,000, and TACO: 2,437), 6,400 instruction-following examples (MAGPIE-IF: 1,400 and RLVR-IFEval: 5,000), and 4,800 general-reasoning examples (BBH: 4,800). All sources are converted into a unified JSON schema containing \texttt{problem}, \texttt{class}, and task-specific verifier metadata. Mathematical and general-reasoning instances provide \texttt{true\_answer}; CODE instances provide input--output \texttt{test\_cases} and, where applicable, \texttt{fn\_name}; IF instances provide \texttt{instruction\_id\_list} and the corresponding verifier arguments.

For final in-distribution (ID) evaluation, we use the held-out test splits of the nine source datasets used for training: GSM8K, MATH, and DAPO-MATH-17k for mathematical reasoning; MBPP, Eurus-2-Code, and TACO for code generation; MAGPIE-IF and RLVR-IFEval for instruction following; and BIG-bench Hard for general reasoning.  These test splits are evaluated separately, and their per-dataset results constitute the ID results in our experiments.

\subsection{Out-of-Distribution Evaluation}

We evaluate out-of-distribution (OOD) generalization on six benchmarks that are excluded from training: IFEval for instruction following, HumanEval for code generation, and AGIEval, ARC-Challenge, LogicBench, and OlympiadBench for general reasoning and mathematical/scientific reasoning.  The OOD protocol retains the trained router, agent adapters, task-appropriate prompt construction, and answer aggregation rule, thereby testing transfer of the learned routing and specialization rather than adaptation to the held-out benchmarks.

% \section{Additional Experiments}
% \subsection{Router Computational Overhead Analysis}
% 11.87ms, 1350.32ms, 154.09ms
% \subsection{Co-Evolutionary Learning Dynamics}
% reward, cumulative update, familiar score
% \subsection{Sensitivity Analysis on Cumulative Threshold}
% 0.7 63.2 72.8 367.77
% 0.6 63.0 72.5 342.22
% 0.5 62.9 71.9 329.06
% 0.8 63.0 72.9 422.12
% 0.9 63.2 72.7 648.01
% \subsection{Ablation on Hidden State Extraction}
% 62.1+72.3 438.12 token

\section{Additional Experiments and Analysis}
\label{app:additional-experiments}

\subsection{Router Computational Overhead Analysis}
\label{app:router-overhead}

A critical requirement for dynamic multi-agent orchestration is that the routing mechanism itself must not become a computational bottleneck. We quantify the efficiency of our predictive familiarity estimator by profiling the average latency per query during test-time inference. All measurements are executed on a single NVIDIA H200 GPU. As detailed in Table~\ref{tab:appendix-overhead}, we compare the router's execution time against the time required for a single agent to generate a complete trajectory, both with and without the vLLM acceleration engine.

\begin{table}[h]
\centering
\small
\resizebox{\linewidth}{!}{\begin{tabular}{l c c}
\toprule
\textbf{Execution Component} & \textbf{Batch Size} & \textbf{Avg. Latency (ms)} \\
\midrule
Predictive Router & 1 & {11.87} \\
\midrule
Completion Generation (Native) & 1 & 1350.32 \\
Completion Generation (vLLM) & 32 & 154.09 \\
\bottomrule
\end{tabular}}
\caption{Computational overhead profiling on a single NVIDIA H200 GPU.}
\label{tab:appendix-overhead}
\end{table}

The empirical results demonstrate that the \sysname~router imposes a negligible computational footprint, averaging merely 11.87\,ms per query. This efficiency arises fundamentally from the architectural design: evaluating familiarity requires extracting intermediate hidden states via a single, partial forward pass. Unlike standard sequence-to-sequence evaluation mechanisms or external LLM-as-a-judge verifiers, our router entirely bypasses the recursive, token-by-token autoregressive decoding phase. Consequently, the routing overhead is less than $1\%$ of the native generation time, ensuring that the system's efficiency gains from activating fewer agents are preserved in end-to-end wall-clock latency.

\subsection{Co-Evolutionary Learning Dynamics}
\label{app:learning-dynamics}

\begin{figure}[h]
  \centering
  \includegraphics[width=.7\columnwidth]{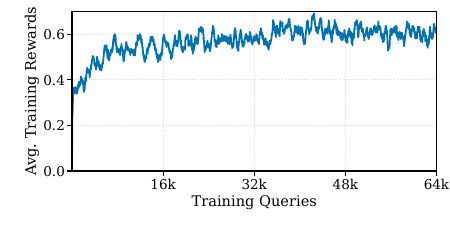}
  \caption{System average reward progression.}
  \label{fig:reward_4000}
\end{figure}

\begin{figure}[h]
  \centering
  \includegraphics[width=.7\columnwidth]{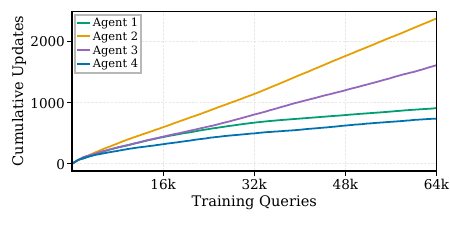}
  \caption{Cumulative updates per agent.}
  \label{fig:update_4000}
\end{figure}

\begin{figure}[h]
  \centering
  \includegraphics[width=.7\columnwidth]{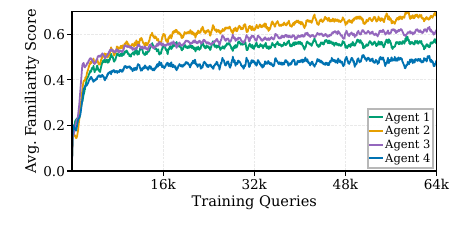}
  \caption{Familiarity score progression.}
  \label{fig:score_4000}
\end{figure}

Tracing the internal training dynamics across the evolutionary timeline illuminates how \sysname~induces capability specialization without human intervention. Figures~\ref{fig:reward_4000}, \ref{fig:update_4000}, and \ref{fig:score_4000} illustrate the progression of system-level rewards, individual agent update frequencies, and the corresponding familiarity scores, respectively.

As shown in Figure~\ref{fig:reward_4000}, the aggregate system reward steadily increases and converges stably, validating the effectiveness of our DAPO--GSPO optimization protocol. Furthermore, Figures~\ref{fig:update_4000} and \ref{fig:score_4000} reveal a clear hierarchical differentiation in agent workloads. Agent 2 rapidly emerges as the primary reasoning engine, executing the highest total volume of updates. Agent 3 specializes strictly in code-generation topologies, maintaining the second-highest update frequency as it digests programming queries. Meanwhile, Agents 1 and 4 secure their niches by supplementing the population on instruction-following, general reasoning, and concise mathematical problems. Importantly, while their overall update frequencies are lower than the primary solvers, their familiarity scores still rise steadily within their specialized semantic clusters. This confirms that our exploration-augmented routing mechanism successfully prevents policy starvation, ensuring all agents continuously evolve and actively contribute to the overarching mixture.

% \subsection{System Scalability on Agent Population}
% \label{app:scalability}

% We verify the scalability of our co-evolutionary framework by expanding the agent population size $N$ within the homogeneous shared-backbone architecture. Separate populations of $N=3$ and $N=6$ agents are trained under identical hyper-parameter configurations.

% \begin{table}[h]
% \centering
% \small
% \resizebox{\linewidth}{!}{\begin{tabular}{c c c c}
% \toprule
% \textbf{Population Size ($N$)} & \textbf{ID Avg.} & \textbf{OOD Avg.} & \textbf{Avg. Tokens} \\
% \midrule
% 3 Agents & XX.X & XX.X & XXX.XX \\
% \rowcolor{gray!15} {4 Agents (Default)} & {63.2} & {72.8} & {367.77} \\
% 6 Agents & XX.X & XX.X & XXX.XX \\
% \bottomrule
% \end{tabular}}
% \caption{Scalability evaluation across different agent population sizes.}
% \label{tab:appendix-scalability}
% \end{table}

% Table~\ref{tab:appendix-scalability} confirms that \sysname~smoothly scales with increased model capacity. Expanding the population to $N=6$ yields fine-grained capability partitioning, allowing the router to cultivate more specialized experts, whereas a smaller pool ($N=3$) enforces broader generalization per agent. This flexibility demonstrates that \sysname~can be seamlessly tuned to match available hardware budgets and task complexity.

\subsection{Hyperparameter Robustness and Sensitivities}
\label{app:hyperparameters}

\paragraph{Cumulative Threshold $\tau$.} 
The cumulative threshold $\tau$ serves as the primary control lever in \sysname, balancing reasoning accuracy against computational cost. Using the default population trained under threshold $0.7$, we conduct a test-time sweep from $\tau=0.5$ to $0.9$ (Table~\ref{tab:appendix-sensitivity-test}). 

\begin{table}[h]
\centering
\small
\begin{tabular}{c c c c}
\toprule
\textbf{Test Threshold ($\tau$)} & \textbf{ID Avg.} & \textbf{OOD Avg.} & \textbf{Avg. Tokens} \\
\midrule
0.5 & 62.9 & 71.9 & 329.06 \\
0.6 & 63.0 & 72.5 & 342.22 \\
\rowcolor{gray!15} {0.7 (Default)} & {63.2} & {72.8} & {367.77} \\
0.8 & 63.0 & 72.9 & 422.12 \\
0.9 & 63.2 & 72.7 & 648.01 \\
\bottomrule
\end{tabular}
\caption{Test-time sensitivity analysis for the default model (trained with $\tau=0.7$).}
\label{tab:appendix-sensitivity-test}
\end{table}

Lowering the test threshold aggressively curbs token expenditure but causes a drop in accuracy, as complex queries fail to trigger sufficient multi-agent consensus. Conversely, elevating it to $0.8$ or $0.9$ triggers massive computational redundancy with little performance gain. Investigating training-time sensitivity involves re-training a population using a stricter threshold of $\tau=0.8$. Testing this population across varying thresholds (Table~\ref{tab:appendix-sensitivity-train}) reveals a substantial performance drop (peak ID: 61.2, OOD: 71.5). A higher training threshold forces over-activation, heavily diluting the targeted specialization feedback and preventing the emergence of sharp domain experts. 

\begin{table}[h]
\centering
\small
\begin{tabular}{c c c c}
\toprule
\textbf{Test Threshold ($\tau$)} & \textbf{ID Avg.} & \textbf{OOD Avg.} & \textbf{Avg. Tokens} \\
\midrule
0.5 & 60.9 & 70.9 & 353.84 \\
0.6 & 61.2 & 71.5 & 371.32 \\
0.7 & 60.8 & 71.4 & 405.36 \\
\rowcolor{gray!15} {0.8 (Train Setting)} & {61.2} & {71.4} & {485.88} \\
0.9 & 61.2 & 71.5 & 676.93 \\
\bottomrule
\end{tabular}
\caption{Evaluation of an alternative model trained with a sub-optimal threshold ($\tau=0.8$).}
\label{tab:appendix-sensitivity-train}
\end{table}

\begin{figure}[h]
  \centering
  \includegraphics[width=.7\columnwidth]{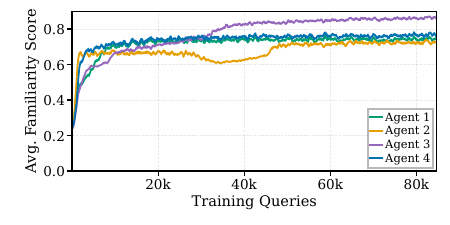}
  \caption{Familiarity score progression under $\lambda=1.0$.}
  \label{fig:score_lambda}
\end{figure}

\paragraph{Familiarity Temperature $\lambda$.}
Examining the sensitivity of the exponential decay parameter $\lambda$ reveals that decreasing the temperature to $\lambda=1.0$ significantly degrades performance (ID: 55.3, OOD: 59.8, Avg. Tokens: 463.72) compared to the default $\lambda=2.0$ (ID: 63.2, OOD: 72.8, Avg. Tokens: 367.77). As illustrated in Figure~\ref{fig:score_lambda}, tracking the average familiarity indicates that the scores under $\lambda=1.0$ artificially inflate and reach the cumulative activation threshold ($\tau=0.7$) far too early in the training process. Meeting the confidence bar so easily causes the router to prematurely halt the subset expansion, severely limiting the exploration-augmented allocation mechanism. This lack of early exploration prevents under-utilized agents from receiving sufficient training queries across diverse semantic domains, ultimately hindering the emergence of well-differentiated specialists.

% \paragraph{Separation Margin $m$.}
% We validate the necessity of the bounded repulsion margin $m$. Increasing the margin to $m=2.0$ drops the accuracy (ID: 59.2, OOD: 70.7) and causes the familiarity score of at least one agent to struggle to grow. More critically, when we completely remove the margin (using pure negative advantage for unbounded repulsion), the familiarity scores of all agents fluctuate violently and fail to grow altogether. This corroborates our geometrical derivation: without the protective margin $m$, unbounded negative gradients induce excessively large shifts in the projection space, permanently collapsing the familiarity representation.

\begin{figure}[h]
  \centering
  \includegraphics[width=.7\columnwidth]{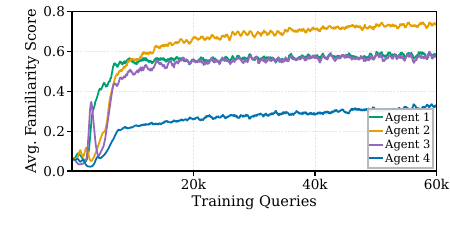}
  \caption{Familiarity score progression under an increased margin ($m=2.0$).}
  \label{fig:score_margin}
\end{figure}

\begin{figure}[h]
  \centering
  \includegraphics[width=.7\columnwidth]{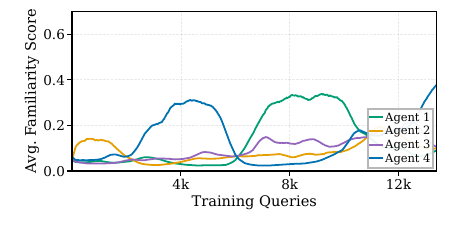}
  \caption{Familiarity score progression without a protective separation margin.}
  \label{fig:score_no_margin}
\end{figure}

\paragraph{Separation Margin $m$.}
Validating the necessity of the bounded repulsion margin $m$ involves testing an increased boundary ($m=2.0$) alongside a completely unbounded variant. Increasing the margin to $m=2.0$ degrades the overall reasoning accuracy (ID: 59.2, OOD: 70.7, Avg. Tokens: 364.74). As depicted in Figure~\ref{fig:score_margin}, an overly aggressive separation margin prevents the familiarity score of at least one agent from growing effectively. More critically, removing the margin (applying $A_{i,q} d_i(q)$ for unbounded repulsion based on negative advantages) results in catastrophic instability. Figure~\ref{fig:score_no_margin} demonstrates that the familiarity scores of all agents fluctuate violently and fail to grow altogether under this unbounded configuration. These observations directly corroborate our geometrical derivation: without the protective margin $m$, unbounded negative gradients induce large shifts in the projection space, collapsing the familiarity representation.

\subsection{Ablation on Routing and Exploration Mechanisms}
\label{app:routing-exploration}

A core premise of \sysname~is that dynamically allocating data based on actual competence coverage is superior to fixed-size routing, and that active exploration is vital for avoiding policy starvation. We justify these architectural choices by ablating the adaptive threshold routing and the exploration bonuses (UCB and historical entropy) during training. The quantitative results of these ablations are summarized in Table~\ref{tab:appendix-ablation-routing}.

\begin{table}[h]
\centering
\small
\resizebox{\linewidth}{!}{\begin{tabular}{l c c c}
\toprule
\textbf{Training Configuration} & \textbf{ID Avg.} & \textbf{OOD Avg.} & \textbf{Avg. Tokens} \\
\midrule
Fixed Top-3 Routing & 63.1 & 72.3 & 817.97 \\
No Exploration (w/o UCB \& Entropy) & 57.1 & 63.9 & 436.60 \\
\rowcolor{gray!15} {Adaptive Threshold (Ours)} & {63.2} & {72.8} & {367.77} \\
\bottomrule
\end{tabular}}
\caption{Ablation of routing allocation and exploration mechanisms.}
\label{tab:appendix-ablation-routing}
\end{table}

\begin{figure}[h]
  \centering
  \includegraphics[width=.7\columnwidth]{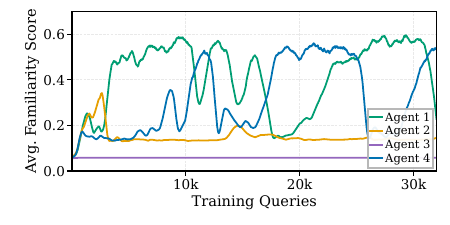}
  \caption{Familiarity score progression under fixed Top-3 training.}
  \label{fig:score_top3}
\end{figure}

\paragraph{Adaptive Threshold vs. Fixed Top-$K$ Training.}
Training an alternative population using a fixed Top-3 routing strategy (while retaining the threshold logic for test-time inference) yields comparable accuracy but severely inflates the token cost to 817.97 tokens per query. Tracing the internal metrics in Figure~\ref{fig:score_top3} explains this inefficiency: forcing the allocation of exactly three agents per query, regardless of the intrinsic difficulty of the prompt, causes the familiarity scores to oscillate violently rather than grow steadily. This instability prevents the system from accurately calibrating its confidence, ultimately failing to route queries efficiently during inference.

\begin{figure}[h]
  \centering
  \includegraphics[width=.7\columnwidth]{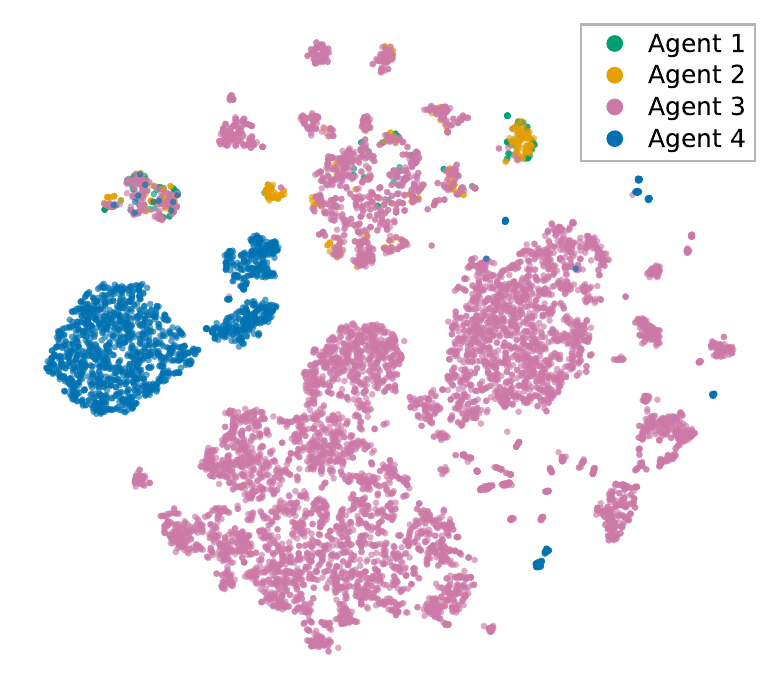}
  \caption{t-SNE visualization of routed hidden states without exploration mechanisms.}
  \label{fig:tsne_no_exp}
\end{figure}

\paragraph{Necessity of Exploration Mechanisms.}
Ablating the UCB exploration bonus and historical entropy regularization during training severely degrades system performance across all metrics (Table~\ref{tab:appendix-ablation-routing}). Visualizing the routed hidden states via t-SNE (Figure~\ref{fig:tsne_no_exp}) confirms that without active exploration, the router suffers from catastrophic mode collapse. Specifically, the vast majority of queries are blindly routed to a single dominant agent (Agent 3), leaving Agents 1 and 2 in complete policy starvation with minimal participation. This failure underscores the absolute necessity of forced exploration for cultivating a balanced, well-differentiated agent population.

\subsection{Ablation on Hidden State Extraction}
\label{app:hidden-state}

A cornerstone of our familiarity estimator is the reliance on mid-layer hidden states rather than the final-layer representations. We empirically justify this architectural design by ablating the feature extraction, forcing the router to predict familiarity strictly from the model's final-layer outputs. 

\begin{table}[h]
\centering
\small
\resizebox{\linewidth}{!}{\begin{tabular}{l c c c}
\toprule
\textbf{Feature Source} & \textbf{ID Avg.} & \textbf{OOD Avg.} & \textbf{Avg. Tokens} \\
\midrule
Final-layer States & 61.1 & 71.3 & 438.12 \\
\rowcolor{gray!15} {Mid-layer States (Ours)} & {63.2} & {72.8} & {367.77} \\
\bottomrule
\end{tabular}}
\caption{Ablation of hidden states extraction depth.}
\label{tab:appendix-ablation-layer}
\end{table}

\begin{figure}[h]
  \centering
  \includegraphics[width=.7\columnwidth]{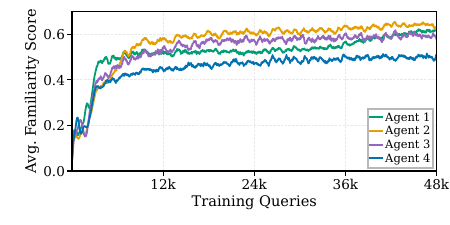}
  \caption{Final-layer familiarity scores progression.}
  \label{fig:final_layer_scores}
\end{figure}

\begin{figure}[h]
  \centering
  \includegraphics[width=.7\columnwidth]{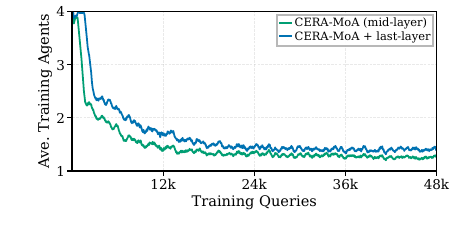}
  \caption{Average training agents activated per query.}
  \label{fig:final_layer_update}
\end{figure}

\begin{figure}[h]
  \centering
  \includegraphics[width=.7\columnwidth]{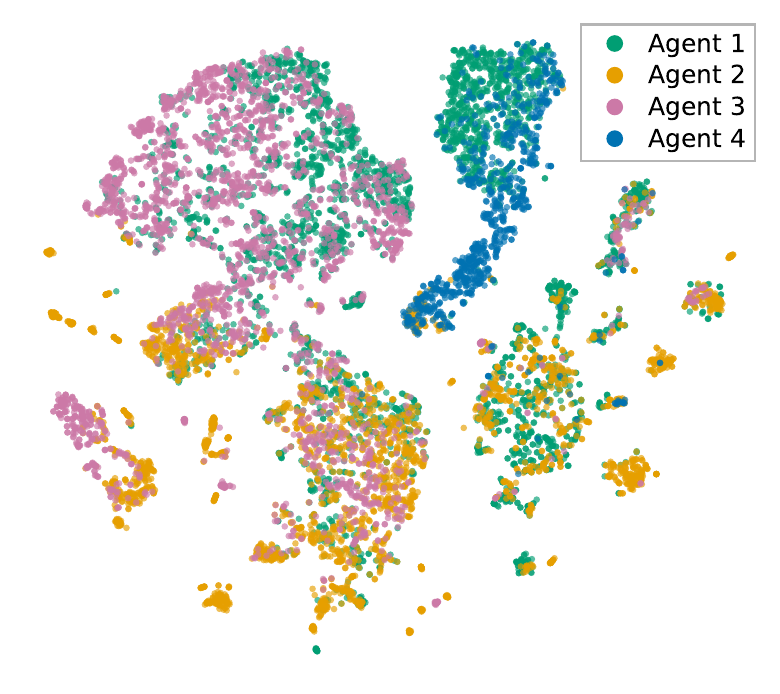}
  \caption{t-SNE visualization of final-layer features.}
  \label{fig:tsne1}
\end{figure}

As presented in Table~\ref{tab:appendix-ablation-layer}, the final-layer variant suffers a notable degradation in both ID and OOD accuracy, accompanied by an unjustified surge in token generation. Figures~\ref{fig:final_layer_scores}, \ref{fig:final_layer_update}, and \ref{fig:tsne1} visualize the internal routing metrics under the final-layer configuration, diagnosing this failure mode.

Figure~\ref{fig:final_layer_scores} demonstrates that final-layer familiarity scores grow slower than their mid-layer counterparts. Consequently, meeting the identical threshold ($\tau=0.7$) during training forces the router to consistently activate a larger subset of agents (Figure~\ref{fig:final_layer_update}), thereby diluting the targeted specialization feedback. Furthermore, the t-SNE visualization (Figure~\ref{fig:tsne1}) reveals that final-layer representations exhibit entanglement, failing to form cleanly separable clusters for distinct agent expertise. We attribute this degradation to the phenomenon investigated by \cite{skean2025layerlayeruncoveringhidden}: while mid-layer hidden states preserve broad, structurally rich semantics, final-layer representations are heavily dominated by immediate, token-specific predictive distributions (autoregressive interference). By extracting features prior to this late-stage output bottleneck, our mid-layer router successfully preserves the representation separability required to map complex queries to the correct domain experts.

\section{Limitations}
\label{app:impact}

% \subsection{Limitations}
% \label{app:limitations}

While \sysname~demonstrates strong adaptability and efficiency across various reasoning benchmarks, our current framework possesses certain analytical limitations. 

First, the predictive familiarity estimator currently evaluates the semantic competence of agents based on the initial user prompt. Consequently, the routing mechanism is inherently optimized for single-turn interactions or fixed-trajectory generation. In long-horizon, multi-turn agentic workflows (e.g., iterative software development or extended multi-agent debate), the required expertise may shift dynamically as the conversational context grows. However, adapting \sysname~to these scenarios is structurally straightforward: it simply entails extending the familiarity evaluation to extract mid-layer hidden states from the complete, updated context window at each interaction step, rather than just the initial instruction. Exploring this continual, step-wise competence tracking remains an actionable and important avenue for future research. 

Second, our current consensus aggregation protocol limits the collaborative potential of the agent population on open-ended generation tasks. While \sysname~successfully leverages familiarity-weighted majority voting for deterministic reasoning (e.g., mathematics), it defaults to outputting the single response from the most familiar agent for open-ended benchmarks, such as code synthesis in MBPP and HumanEval. Although this design choice reduces computational overhead, it inherently bypasses the advantages of multi-agent generation. In these generative scenarios, synthesizing diverse candidate solutions from multiple activated experts could potentially yield a more robust and optimal final output than relying on a single specialist. To address this, a promising extension is the integration of a generative aggregator, such as an LLM-based meta-thinker, to dynamically synthesize the completions of the selected expert subset. Importantly, this advanced answer aggregation approach is architecturally fully compatible with our current design.

\end{document}